\documentclass[10pt,twocolumn,letterpaper]{article}

\usepackage{cvpr}              
\usepackage{chen}
\usepackage{wrapfig}
\usepackage{arydshln}
\newcommand{\cmark}{\ding{51}}%
\newcommand{\xmark}{\ding{55}}%
\usepackage[pagebackref,breaklinks,colorlinks]{hyperref}
\usepackage{color, soul}
\setulcolor{blue}
\usepackage{graphicx}
\definecolor{mygray2}{gray}{.6}
\definecolor{mywarning}{RGB}{233,144,61}
\definecolor{red}{RGB}{255,0,0}

\def\cvprPaperID{3286}
\def\confName{CVPR}
\def\confYear{2023}

\begin{document}

\title{Boosting Video Object Segmentation via Space-time Correspondence Learning}

\renewcommand{\thefootnote}{\fnsymbol{footnote}}

\author{%
Yurong Zhang\textsuperscript{1}\footnotemark[1] , Liulei Li\textsuperscript{2}\footnotemark[1] , Wenguan Wang\textsuperscript{2}\footnotemark[2] , Rong Xie\textsuperscript{1}, Li Song\textsuperscript{1}, Wenjun Zhang\textsuperscript{1} \\
\small \textsuperscript{1}School of Electronic Information and Electrical Engineering, Shanghai Jiao Tong University ~\textsuperscript{2}ReLER, CCAI, Zhejiang University \\
\small \url{https://github.com/wenguanwang/VOS_Correspondence}
}

\maketitle

\begin{abstract}
\footnotetext[1]{The first two authors contribute equally to this work.}\footnotetext[2]{Corresponding author.}
Current top-leading solutions for video object segmentation (VOS) typically follow a \textbf{matching-based} regime:~for each$_{\!}$ query$_{\!}$ frame,$_{\!}$ the$_{\!}$ segmentation$_{\!}$ mask$_{\!}$ is$_{\!}$ inferred$_{\!}$ accor- ding to its correspondence to previously processed and~the first$_{\!}$ annotated$_{\!}$ frames.$_{\!}$ They$_{\!}$ simply$_{\!}$ exploit$_{\!}$ the$_{\!}$ supervisory signals from the groundtruth masks for learning mask pre- diction$_{\!}$ only,$_{\!}$ without$_{\!}$ posing$_{\!}$ any$_{\!}$ constraint$_{\!}$ on$_{\!}$ the$_{\!}$ space-time correspondence$_{\!}$ matching,$_{\!}$ which,$_{\!}$ however,$_{\!}$ is$_{\!}$ the$_{\!}$ fundamen-
 tal$_{\!}$ building$_{\!}$ block$_{\!}$ of$_{\!}$ such$_{\!}$ regime.$_{\!}$ To$_{\!}$ alleviate$_{\!}$ this$_{\!}$ crucial$_{\!}$~yet commonly$_{\!}$ ignored$_{\!}$ issue,$_{\!}$ we$_{\!}$ devise$_{\!}$ a$_{\!}$ correspondence-aware training$_{\!}$ framework,$_{\!}$ which$_{\!}$ boosts$_{\!}$ matching-based$_{\!}$ VOS$_{\!}$ so- lutions$_{\!}$ by$_{\!}$ explicitly$_{\!}$ encouraging$_{\!}$ robust$_{\!}$ correspondence$_{\!}$~ma- tching during network learning. Through comprehensively exploring the intrinsic coherence in videos on pixel and ob- ject levels, our algorithm reinforces the standard, fully~su- pervised training$_{\!}$ of$_{\!}$ mask$_{\!}$ segmentation$_{\!}$ with$_{\!}$ label-free, con- trastive$_{\!}$ correspondence$_{\!}$ learning.$_{\!}$ Without$_{\!}$ neither$_{\!}$ requiring$_{\!}$  extra annotation cost during training, nor causing speed delay during deployment, nor incurring architectural modification, our algorithm provides solid performance gains on four widely used benchmarks, \ie, DAVIS2016\&2017, and YouTube-VOS2018\&2019, on the top of famous matching-based VOS solutions.


\end{abstract}

\section{Introduction}
	\vspace{-3pt}
In this work, we address the task of (one-shot) video~ob- ject$_{\!}$ segmentation$_{\!}$ (VOS)$_{\!}$~\cite{caelles2017one,wang2018semi,wang2021survey}.$_{\!}$ Given$_{\!}$ an$_{\!}$ input$_{\!}$ video$_{\!}$ with groundtruth object masks in the first frame, VOS aims at accurately segmenting the annotated objects in the subsequent frames.  As one of the most challenging tasks in computer vision, VOS benefits a wide range of applications including augmented reality and interactive video editing$_{\!}$~\cite{wang2017selective}.
\begin{figure}[t]

	\begin{center}
		\includegraphics[width=1.\linewidth]{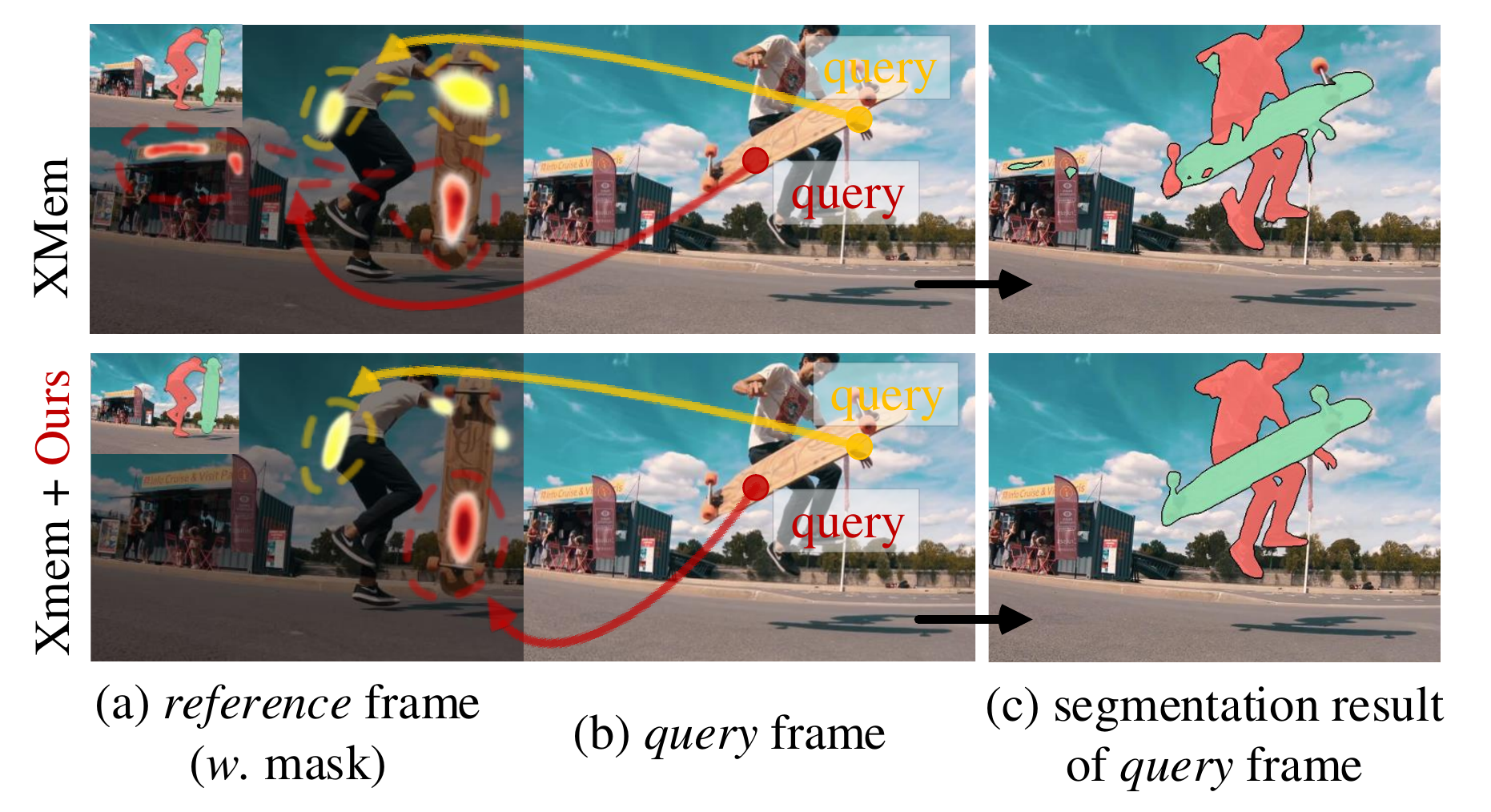}
		\put(-236,107){\scriptsize \rotatebox{90}{\cite{cheng2022xmem}}}
		\end{center}
	\vspace{-18pt}
	\captionsetup{font=small}
	\caption{\small $_{\!\!\!}$(a-b)$_{\!}$ shows$_{\!}$ some$_{\!}$ correspondences$_{\!}$ between$_{\!}$ a$_{\!}$ reference$_{\!}$ frame$_{\!}$ and$_{\!}$ a$_{\!}$ query$_{\!}$ frame.$_{\!}$ (c)$_{\!}$ gives$_{\!}$ mask$_{\!}$ prediction.$_{\!}$  XMem$_{\!}$~\cite{cheng2022xmem},$_{\!}$ even a top-leading matching-based VOS solution, still suffers from unreliable correspondence. In contrast, with our correspondence-aware training strategy, robust space-time correspondence can be established, hence leading to better mask-tracking results.
	}
	\label{fig:1}
	\vspace{-15pt}
\end{figure}

Modern VOS solutions are built upon fully supervised deep learning techniques and the top-performing ones$_{\!}$~\cite{cheng2021rethinking,cheng2022xmem} largely follow a \textit{matching-based} paradigm, where the object masks for a new coming frame (\ie, query frame) are generated according to the correlations between the query frame and the previously segmented as well as first annotated frames (\ie, reference frames), which are stored~in~an outside$_{\!}$ memory.$_{\!}$ It$_{\!}$ is$_{\!}$ thus$_{\!}$ apparent$_{\!}$ that$_{\!}$ the$_{\!}$ module$_{\!}$ for$_{\!}$~cross-$_{\!}$ frame$_{\!}$ matching$_{\!}$ (\ie,$_{\!}$ space-time$_{\!}$ correspondence$_{\!}$ modeling) plays the central role in these advanced VOS systems. Nevertheless, these matching-based solutions are simply trained under the direct supervision of the groundtruth segmentation masks. In other words, during training, the whole VOS system is purely optimized towards accurate segmentation mask prediction, yet without taking into account any \textit{explicit} constraint/regularization on the central component --- space-time correspondence matching. This comes with a legitimate concern for sub-optimal performance, since there is no any solid guarantee of truly establishing reliable cross-frame correspondence during network learning.$_{\!}$ Fig.$_{\!}$~\ref{fig:1}(a)$_{\!}$~offers a visual evidence for this viewpoint. XMem$_{\!}$~\cite{cheng2022xmem}, the latest state-of-the-art matching-based VOS solution, tends to struggle at discovering valid space-time correspondence;
indeed,$_{\!}$ some$_{\!}$ background$_{\!}$ pixels/patches$_{\!}$ are$_{\!}$ incorrectly$_{\!}$ reco- gnized as highly correlated to the query foreground.



The aforementioned discussions motivate us to propose a new, space-time correspondence-aware training framework which addresses the weakness of existing matching-based VOS solutions in an elegant and targeted manner. The core\\
\noindent idea$_{\!}$ is$_{\!}$ to$_{\!}$ empower$_{\!}$ the$_{\!}$ matching-based$_{\!}$ solutions$_{\!}$ with$_{\!}$~en- hanced$_{\!}$ robustness$_{\!}$ of$_{\!}$ correspondence$_{\!}$ matching,$_{\!}$ through$_{\!}$~mi- ning complementary yet \textit{free} supervisory signals from the inherent$_{\!}$ nature$_{\!}$ of$_{\!}$ space-time$_{\!}$ continuity$_{\!}$ of$_{\!}$ training$_{\!}$ video$_{\!}$ sequences.$_{\!}$ In more detail,$_{\!}$ we$_{\!}$ comprehensively$_{\!}$ investigate the$_{\!}$ coherence$_{\!}$ nature$_{\!}$ of$_{\!}$ videos$_{\!}$ on$_{\!}$ both$_{\!}$ pixel$_{\!}$ and$_{\!}$ object$_{\!}$ levels:  \textbf{i)}$_{\!}$
\textit{pixel-level$_{\!}$ consistency}:$_{\!}$ spatiotemporally$_{\!}$ proximate$_{\!}$ pixels/patches tend$_{\!}$ to$_{\!}$ be$_{\!}$ consistent;$_{\!}$ and$_{\!}$ \textbf{ii)}$_{\!}$ \textit{object-level$_{\!}$ coherence}:$_{\!}$ visual$_{\!}$ semantics$_{\!}$ of$_{\!}$ same object instances at different timesteps$_{\!}$ tend to$_{\!}$ {retain unchanged}.$_{\!}$ By$_{\!}$ accommodating these two properties to an unsupervised learning scheme, we$_{\!}$ give$_{\!}$ more explicit direction on the correspondence matching process, hence promoting the VOS model to learn dense discriminative and object-coherent visual representation for robust, matching-based mask tracking (see Fig.$_{\!}$~\ref{fig:1}~(b-c)).

It is worth mentioning that, beyond boosting the segmentation performance, our space-time correspondence-aware training framework enjoys several compelling facets.~\textbf{First}, our algorithm supplements the standard, fully supervised training$_{\!}$ paradigm$_{\!}$ of$_{\!}$ matching-based$_{\!}$ VOS$_{\!}$ with$_{\!}$ \textit{self-training} of$_{\!}$ space-time$_{\!}$ correspondence.$_{\!}$ {As$_{\!}$ a$_{\!}$ result, it does not cause any extra annotation burden.} 
\textbf{Second}, our algorithm is fully compatible with current popular matching-based VOS solu- tions$_{\!}$~\cite{cheng2021rethinking,cheng2022xmem}, without particular adaption to the segmenta- tion$_{\!}$ network$_{\!}$ architecture.$_{\!}$ This$_{\!}$ is$_{\!}$ because$_{\!}$ the$_{\!}$ learning$_{\!}$~of$_{\!}$ the correspondence$_{\!}$ matching$_{\!}$ only$_{\!}$ happens$_{\!}$ in$_{\!}$ the$_{\!}$ visual$_{\!}$ embedding space.  \textbf{Third}, {as a training framework, our algorithm does not produce additional computational budget to the applied VOS models} during the deployment phase.

$_{\!}$We$_{\!}$ make$_{\!}$ extensive$_{\!}$ experiments$_{\!}$ on$_{\!}$ various$_{\!}$ gold-standard$_{\!}$ VOS datasets, \ie, DAVIS2016\&2017$_{\!}$~\cite{pont2017davis}, and YouTube-VOS2018\&2019$_{\!}$~\cite{xu2018youtube}.$_{\!}$ We$_{\!}$ empirically$_{\!}$ prove$_{\!}$ that, on the top of recent matching-based VOS models, \ie, STCN$_{\!}$~\cite{cheng2021rethinking} and
XMem$_{\!}$~\cite{cheng2022xmem}, our approach gains impressive results, surpassing all$_{\!}$ existing$_{\!}$ state-of-the-arts.$_{\!}$ Concretely,$_{\!}$ in$_{\!}$ multi-object$_{\!}$ scenarios,$_{\!}$ it$_{\!}$ improves$_{\!}$ STCN$_{\!}$ by$_{\!}$ \textbf{1.2}\%,$_{\!}$ \textbf{2.3}\%,$_{\!}$ and$_{\!}$ \textbf{2.3}\%,$_{\!}$~and XMem$_{\!}$ by$_{\!}$ \textbf{1.5}\%,$_{\!}$ \textbf{1.2}\%,$_{\!}$ and$_{\!}$ \textbf{1.1}\%$_{\!}$ on$_{\!}$ DAVIS2017$_{\!val}$,$_{\!\!}$ Youtube-
VOS2018$_{val}$,$_{\!}$ as$_{\!}$ well$_{\!}$ as$_{\!}$ Youtube-VOS2019$_{val}$,$_{\!}$ respectively, in$_{\!}$ terms$_{\!}$ of$_{\!}$  $J\&F$.$_{\!}$ {Besides},$_{\!}$ it$_{\!}$ respectively$_{\!}$  promotes$_{\!}$  STCN and
XMem by \textbf{0.4}\% and \textbf{0.7}\% on single-object benchmark dataset$_{\!}$ DAVIS2016$_{val}$.

\section{Related Work}

\noindent\textbf{(One-Shot) Video Object Segmentation.} Recent VOS solutions can be roughly categorized into three groups: \textbf{i)} \textit{Online learning} based methods adopt online fine-tuning~\cite{caelles2017one,voigtlaender2017online,maninis2018video}$_{\!}$ or adaption~\cite{meinhardt2020make,bhat2020learning,robinson2020learning,park2021learning}$_{\!}$ techniques to accommodate a pre-trained generic segmentation network to the test-time target objects. Though impressive, they
are typically hyper- parameter sensitive and {low efficient}. \textbf{ii)} \textit{Propagation-based}\\
\noindent methods~\cite{wen2015jots,marki2016bilateral,perazzi2017learning,jang2017online,hu2017maskrnn,cheng2018fast,oh2018fast,hu2018motion,bao2018cnn,zhang2019fast,wang2019fast,ventura2019rvos,chen2020state}~formu- late$_{\!}$ VOS$_{\!}$ as$_{\!}$ a$_{\!}$ frame-by-frame$_{\!}$ mask$_{\!}$ propagation$_{\!}$ process. Though compact, they heavily rely on the previous segmentation$_{\!}$ mask,$_{\!}$ hence$_{\!}$ easily$_{\!}$ trapping$_{\!}$ in$_{\!}$ occlusion$_{\!}$ cases$_{\!}$ and$_{\!}$~{suffering from} error accumulation. \textbf{iii)} \textit{Matching-based methods} instead leverage the first annotated frame (and~previous segmented frames) to build an explicit object~model, according  to$_{\!}$ which$_{\!}$ query$_{\!}$ pixels$_{\!}$ are$_{\!}$ matched$_{\!}$ and$_{\!}$ classified \cite{chen2018blazingly,voigtlaender2019feelvos,yang2021collaborative}. As a landmark in this line, STM$_{\!}$~\cite{oh2019video} introduces an external memory for explicitly and persistently storing the~representations and masks of past frames, allowing for long-term matching. Since then, matching-based solutions \cite{seong2020kernelized,lu2020video,miao2020memory,cheng2021modular,hu2021learning,xie2021efficient,wang2021swiftnet,seong2021hierarchical,mao2021joint,cheng2021rethinking, park2022per, li2022recurrent} dominate this area due to the superior performance and high efficiency~\cite{wang2021survey}.


Recent studies for matching-based VOS mainly focus on improving network designs, through, for instance, building more efficient memory$_{\!}$~\cite{lu2020video,wu2020memory,li2020fast,liang2020video,cheng2021rethinking,cheng2022xmem}, adopting local matching$_{\!}$~\cite{seong2020kernelized,hu2021learning,yu2022batman}, and incorporating background con-
text$_{\!}$~\cite{yang2021collaborative}. However, our contribution is orthogonal to these studies, as we advance the matching-based regime in the~as- pect$_{\!}$ of$_{\!}$ model$_{\!}$ learning. {$_{\!}$ We$_{\!}$ devise$_{\!}$ a$_{\!}$ new$_{\!}$ training$_{\!}$ framework that improves the standard, supervised segmentation training protocol with self-constrained correspondence learning. We show our algorithm can be seamlessly incorporated into the latest arts$_{\!}$~\cite{cheng2021rethinking,cheng2022xmem} with notable performance gains.

\noindent\textbf{Self-supervised Space-time Correspondence Learning.} Capturing cross-frame correlations is a long-standing task in the {field} of computer vision, due to its vital role in many video applications such as optical flow estimation, and ob- ject$_{\!}$ tracking.$_{\!}$ A$_{\!}$ line$_{\!}$ of$_{\!}$ recent$_{\!}$ work$_{\!}$ tackles$_{\!}$  this$_{\!}$  problem$_{\!}$  in$_{\!}$  a
 self-supervised$_{\!}$ learning$_{\!}$
fashion.$_{\!}$ The$_{\!}$ methods$_{\!}$ can$_{\!}$ be$_{\!}$ divided
 into three classes: \textbf{i)} \textit{Reconstruction} based methods enforce the network to reconstruct a query frame from a neighboring frame$_{\!}$~\cite{vondrick2018tracking,lai2019self,lai2020mast,wang2021contrastive,jeon2021mining,araslanov2021dense,bhat2020learning,li2022locality}, so as to find accurate alignment between the query and reference frames. \textbf{ii)} \textit{Cycle-consistency} based methods$_{\!}$~\cite{wang2019unsupervised,wang2019learning,li2019joint,lu2020learning,jabri2020space,zhao2021modelling,son2022contrastive}     conduct forward-backward tracking and learn correspondence by penalizing the disagreement between start and end points. \textbf{iii)} \textit{Contrastive learning} based methods \cite{jeon2021mining,xu2021rethinking,kim2020adversarial,araslanov2021dense,sharma2022mvdecor}$_{\!}$ emerged very recently, inspired by the astonishing success of contrastive learning in self-supervised image representation learning. Their core idea is to distinguish confident correspondences from a large set of unlikely ones.

Although$_{\!}$ a$_{\!}$ few$_{\!}$ correspondence$_{\!}$ learning$_{\!}$ methods$_{\!}$ also$_{\!}$~report performance on mask-tracking, they {confine focus to} the self-supervised setting and simply treat VOS as an exemplar application task without contributing neither dedicated model design nor specific insight to VOS. This work represents a very early (if not the first) effort towards boosting supervised learning of VOS with self-supervised corres- pondence learning, within a principled training framework. 
Hence our ultimate goal is supervised learning of VOS, yet space-time correspondence learning is a mean to this end.

\noindent\textbf{Object-level Self-supervised Learning.} Self-supervised visual representation learning aims to learn transferable$_{\!}$~fea- tures$_{\!}$ with$_{\!}$ massive$_{\!}$ unlabeled$_{\!}$ data.$_{\!}$ Recently,$_{\!}$ contrastive$_{\!}$~lear- ning based methods$_{\!}$~\cite{chen2020simple, he2020momentum,dwibedi2021little,zhu2021improving,grill2020bootstrap,zbontar2021barlow,wu2018unsupervised,tian2020makes,xie2021detco} made$_{\!}$ rapid$_{\!}$ progress. $_{\!\!}$ They$_{\!}$ are$_{\!}$ build$_{\!}$ upon$_{\!}$ an$_{\!}$ \textit{instance$_{\!}$ discrimination$_{\!}$} task that maximizes the agreement between different augmented views$_{\!}$ of$_{\!}$ the$_{\!}$ same$_{\!}$ image.$_{\!}$ Yet,$_{\!}$ it$_{\!}$ then$_{\!}$ became$_{\!}$ apparent$_{\!}$ that image-level pretraining is suboptimal to dense prediction$_{\!}$ tasks$_{\!}$~\cite{wang2021dense,wei2021aligning}, due to the discrepancy between holistic repre- sentation$_{\!}$ and$_{\!}$ fine-grained$_{\!}$ task$_{\!}$ nature.$_{\!}$ Hence$_{\!}$ a$_{\!}$ growing$_{\!}$ num-$_{\!\!}$ ber$_{\!}$ of$_{\!}$ work$_{\!}$ investigate$_{\!}$ pixel-level$_{\!}$ pretraining$_{\!}$~\cite{araslanov2021dense, bian2022learning, wang2021exploring,xie2021propagate, wang2021dense,o2020unsupervised}.$_{\!}$ Though$_{\!}$ addressing$_{\!}$ local$_{\!}$ semantics,$_{\!}$ they$_{\!}$~fail$_{\!}$ to$_{\!}$ learn object-level visual properties. In view of the limitations of image-level and pixel-level self-supervised learning, some latest efforts$_{\!}$~\cite{henaff2021efficient,wei2021aligning,henaff2022object,li2022contextual,xie2021unsupervised,yin2022proposal,selvaraju2021casting,van2021unsupervised} turn to exploring object-level pretraining, with the aid of heuristic object proposals$_{\!}$~\cite{wei2021aligning,xie2021unsupervised,henaff2021efficient}, saliency$_{\!}$~\cite{selvaraju2021casting,van2021unsupervised}, or clustering$_{\!}$~\cite{henaff2022object}.
We assimilate the insight of object-level representation learning and perform adaption to self-supervised correspondence learning. This leads to a comprehensive solution for both dense and object-oriented correlation modeling, hence grasping the central properties of matching-based VOS.

\section{Methodology}


Fig.$_{\!}$~\ref{fig:3} depicts$_{\!}$ a$_{\!}$ diagram$_{\!}$ of$_{\!}$ our$_{\!}$ algorithm.$_{\!}$ Before$_{\!}$ elucida- ting our correspondence-aware training framework for$_{\!}$ VOS (\textit{cf}.$_{\!}$~\S\ref{sec:3.2}), we first formalize the task of interest and provide preliminaries on recent advanced matching-based$_{\!}$ VOS$_{\!}$ so- lutions$_{\!}$ (\textit{cf}.$_{\!}$~\S\ref{sec:3.1}).$_{\!}$ Finally,$_{\!}$ \S\ref{sec:3.3} gives implementation details.

\subsection{Problem Statement and Preliminaries}\label{sec:3.1}

\noindent\textbf{Task$_{\!}$ Setup.$_{\!}$} In$_{\!}$ VOS,$_{\!}$ the$_{\!}$ target$_{\!}$ object$_{\!}$ is$_{\!}$ predefined by$_{\!}$ a$_{\!}$ re- ference$_{\!}$ mask$_{\!}$ in$_{\!}$ the$_{\!}$ first$_{\!}$ frame.$_{\!}$ Formally,$_{\!}$ given$_{\!}$ a$_{\!}$ video with$_{\!}$ $T_{\!}$ frames$_{\!}$ $\mathcal{I}_{\!}\!=_{\!}\!\{I_t\}_{t=1\!\!}^{T}$ and$_{\!}$ the$_{\!}$ first-frame$_{\!}$ reference$_{\!}$ mask$_{\!}$~$Y_1$,$_{\!\!}$ a~robust VOS solution $f$ should exploit~$Y_1$~to produce accu- rate$_{\!}$ object$_{\!}$ masks$_{\!}$ $\{\hat{Y}_t\}_{t=2}^{T}$ for$_{\!}$ the$_{\!}$ rest$_{\!}$ $T\!-\!1_{\!}$ frames$_{\!}$ $\{I_t\}_{t=2}^{T}$:
\vspace{-4pt}
 \begin{equation}\small\label{eq:1}
   \{\hat{Y}_t\}_{t=2}^{T} = f(\{I_t\}_{t=1}^{T}, Y_1).
 \vspace{-3pt}
 \end{equation}
Modern$_{\!}$ VOS$_{\!}$ solutions$_{\!}$ often$_{\!}$ rely$_{\!}$ on$_{\!}$ supervised$_{\!}$ deep$_{\!}$~learn- ing techniques.$_{\!}$ For a training video $\mathcal{I}_{\!}\!=_{\!}\!\{I_t\}_{t=1}^{T}$, let us~de- note$_{\!}$ its$_{\!}$ groundtruth$_{\!}$ mask$_{\!}$ sequence$_{\!}$ as$_{\!}$ $\mathcal{Y}_{\!}\!=_{\!}\!\{Y_{t}\}_{t=1}^{T}$.$_{\!}$~The$_{\!}$~op- timal$_{\!}$ solution$_{\!}$ $f^*_{\!}$ is$_{\!}$ found$_{\!}$ by$_{\!}$ minimizing$_{\!}$ a$_{\!}$ supervised$_{\!}$ seg-$_{\!}$ mentation$_{\!}$ loss$_{\!}$ $\mathcal{L}_{\text{SEG}}$,$_{\!}$ on$_{\!}$ a$_{\!}$ set$_{\!}$ of$_{\!}$ $N_{\!}$ training$_{\!}$ pairs$_{\!}$ $\{\mathcal{I}_n,_{\!} \mathcal{Y}_{n\!}\}^N_{n=1}$:$_{\!\!}$
\vspace{-4pt}
 \begin{equation}\small\label{eq:2}
f^* = \mathop{\arg\min}_{f} \frac{1}{NT}\sum\nolimits_n\sum\nolimits_t\mathcal{L}_{\text{SEG}}(\hat{Y}_t,Y_{t}),
 \vspace{-3pt}
 \end{equation}
where $\mathcal{L}_{\text{SEG}}$ is typically the well-known cross-entropy loss.

\begin{figure}[t]
	\begin{center}
		\includegraphics[width=1.\linewidth]{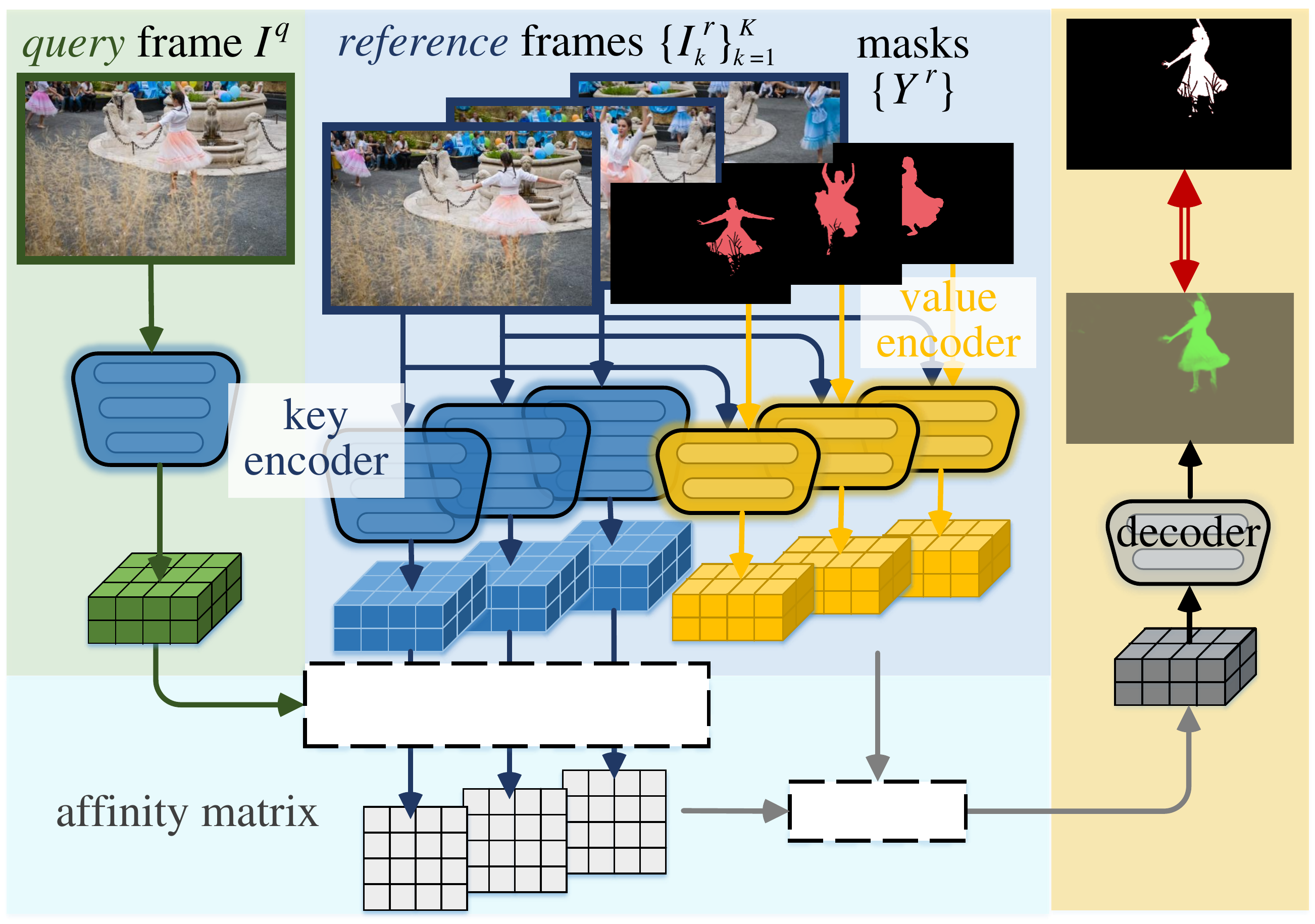}
		\put(-212,90){\large $\kappa$}
		\put(-167,75){\large $\kappa$}
		\put(-150,80){\large $\kappa$}
		\put(-131,83){\large $\kappa$}
		\put(-108,78){\large $\upsilon$}
		\put(-88,82){\large $\upsilon$}
		\put(-70,86){\large $\upsilon$}
		\put(-235,53){\scriptsize $\bm{K}^{q}$}
		\put(-190,50){\scriptsize $\bm{K}^{r}$}
		\put(-54,62){\scriptsize $\bm{V}^{r}$}
		\put(-13,38){\scriptsize $\bm{V}^{q}$}
		\put(-20,125){\scriptsize $\mathcal{L}_{\text{SEG}}$}
		\put(-20,118){\tiny ($_{\!}$~Eq.\ref{eq:2})}
		\put(-208,8){\scriptsize $A$}
		\put(-181,38){\scalebox{.48} {$A(i,j)\!=\!\frac{\textstyle\exp\big(\langle\bm{K}^{r\!}(i), \bm{K}^{q\!}(j)\rangle\big)}{\textstyle\sum\nolimits_{i'}\exp\big(\langle\bm{K}^{r\!}({i'}), \bm{K}^{q\!}(j)\rangle\big)}$}}
		\put(-94,18){\scalebox{.65} {$\bm{V}^{q}=\bm{V}^{r}\!A$}}
		\end{center}
	\vspace{-18pt}
	\captionsetup{font=small}
	\caption{Illustration$_{\!}$ of$_{\!}$ recent$_{\!}$ advanced$_{\!}$ matching-based$_{\!}$ VOS$_{\!}$ so- lutions$_{\!}$~\cite{cheng2021rethinking,cheng2022xmem}, which are simply trained with the manually annotated segmentation masks, without posing any explicit supervisory signal on correspondence matching (\ie, affinity estimation).
	}
	\label{fig:2}
	\vspace{-10pt}
\end{figure}
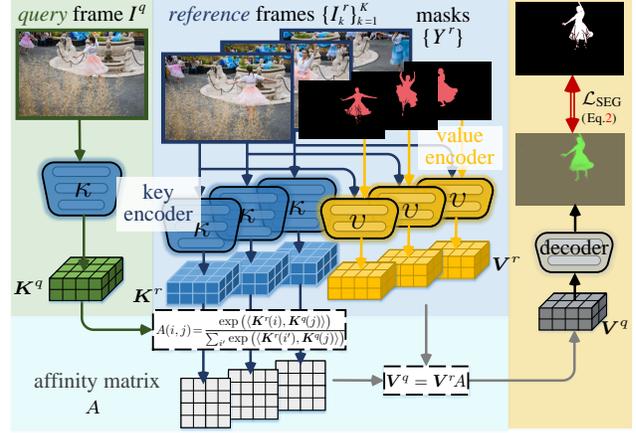

\noindent\textbf{Revisit$_{\!}$ Matching-based$_{\!}$ VOS$_{\!}$ Solutions.$_{\!}$} Since$_{\!}$ the$_{\!}$ seminal$_{\!}$ work$_{\!}$ of$_{\!}$ STM$_{\!}$~\cite{oh2019video},$_{\!}$ VOS$_{\!}$ solutions$_{\!}$ largely$_{\!}$ adopt a$_{\!}$ matching-based$_{\!}$ regime,$_{\!}$ where$_{\!}$ the$_{\!}$ mask$_{\!}$ $\hat{Y}^{_{\!}q\!}$ of$_{\!}$ current$_{\!}$ query$_{\!}$ frame$_{\!}$ ${I}^{q\!}$ is$_{\!}$~predicted$_{\!}$ according$_{\!}$ to$_{\!}$ the$_{\!}$ correlations$_{\!}$ between$_{\!}$ ${I}^{q\!}$ and$_{\!}$~past processed frames $\{I^r_k\}_{k=1}^{K}$. The reference frames $\{I^r_k\}_{k}$  and masks $\{\hat{Y}^r_k\}_{k}$ are stored~in an$_{\!}$ external$_{\!}$ \textit{memory},$_{\!}$ easing$_{\!}$ the$_{\!}$  access$_{\!}$ of$_{\!}$ long-term$_{\!}$ historic$_{\!}$ context.$_{\!\!}$ We$_{\!}$ are$_{\!}$ particularly$_{\!}$~in-
terested$_{\!}$ in$_{\!}$ the$_{\!}$ latest$_{\!}$ matching-based$_{\!}$ models,$_{\!}$ \ie,$_{\!}$ STCN$_{\!}$~\cite{cheng2021rethinking} and XMem$_{\!}$~\cite{cheng2022xmem}, due to their high performance and elegant model design. Specifically, they have two core components:
\begin{itemize}[leftmargin=*]
	\setlength{\itemsep}{0pt}
	\setlength{\parsep}{-2pt}
	\setlength{\parskip}{-0pt}
	\setlength{\leftmargin}{-10pt}
	\vspace{-4pt}
	\item \textit{Key$_{\!}$ encoder$_{\!}$} $\kappa$ takes$_{\!}$ one$_{\!}$ single$_{\!}$ frame$_{\!}$ image$_{\!}$ as$_{\!}$ input and outputs a dense visual feature, \ie, $\bm{K}\!=\!\kappa(I)\!\in\!\mathbb{R}^{C\!\times\! HW\!}$, with $HW$ spatial dimension and $C$ channels.
	
	\item \textit{Value encoder} $\upsilon$ extracts mask-embedded representation for paired frame and mask, \ie, $\bm{V}\!=\!\upsilon(I, \hat{Y})\!\in\!\mathbb{R}^{D_{\!}\times\! HW\!}$.

	\vspace{-4pt}
\end{itemize}
Given$_{\!}$ a$_{\!}$ query$_{\!}$ frame$_{\!}$ ${I}^{q\!}$ and$_{\!}$ $K_{\!}$ reference$_{\!}$ frame$_{\!}$ and$_{\!}$ mask$_{\!}$ pairs$_{\!}$~$\{(I^r_k, \hat{Y}^r_k)\}^K_{k=1\!}$ stored in the memory, we have: {query key} $\bm{K}^{q\!}\!\in\!\mathbb{R}^{C\!\times\! HW\!}$, {memory key}~$\bm{K}^{r\!}\!\in\!\mathbb{R}^{C\!\times\! KHW\!}$,$_{\!}$ and$_{\!}$ memory value$_{\!}$ $\bm{V}^{r\!}\!\in_{\!}\!\mathbb{R}^{D\!\times\! KHW\!}$.$_{\!}$ Then$_{\!}$ we$_{\!}$ compute$_{\!}$ the$_{\!}$ (augmented)$_{\!}$  affinity matrix $A\!\in\![0,1]^{KHW\times HW}$ between $\bm{K}^r$ and $\bm{K}^q$:
\vspace{-3pt}
\begin{equation}\small
    \begin{aligned}\label{eq:3}
A(i,j)\!=\!\frac{\exp\big(\langle\bm{K}^{r\!}(i), \bm{K}^{q\!}(j)\rangle\big)}{\textstyle\sum\nolimits_{i'}\exp\big(\langle\bm{K}^{r\!}({i'}), \bm{K}^{q\!}(j)\rangle\big)}, 
    \end{aligned}
    \vspace{-3pt}
\end{equation}
where $\bm{K}(i)\!\in\!\mathbb{R}^{C\!}$ denotes the feature vector~of $i$-th$_{\!}$ position$_{\!}$ in$_{\!}$ $\bm{K}$,$_{\!}$ and$_{\!}$ $\langle\cdot, \cdot\rangle$$_{\!}$ is a similarity measure, \eg, $\ell_2$ distance. In this way, $A(i,j)\!\in\![0,1]$ -- the $(i,j)$-th element in the normalized affinity $A$ -- signifies the proximity between $i$-th$_{\!}$ pixel in the reference $\{I^r_k\}_{k}$ and $j$-th$_{\!}$ pixel in the query ${I}^{q}$.

\begin{figure*}[t]
    \vspace{-5pt}
	\begin{center}
		\includegraphics[width=1.\linewidth]{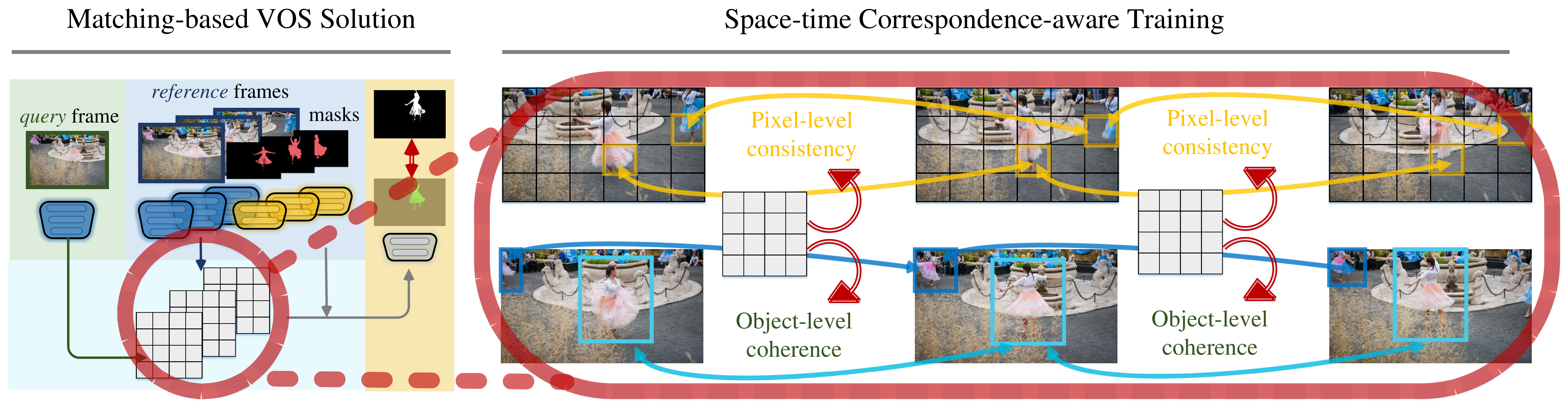}
		\put(-92,43){\scriptsize $\mathcal{L}_{\text{OCL}}$}
		\put(-91.5,36){\tiny (Eq.$_{\!}$~\ref{eq:11})}
		\put(-92,63){\scriptsize $\mathcal{L}_{\text{PCL}}$}
		\put(-90,56){\tiny (Eq.$_{\!}$~\ref{eq:7})}
		\put(-223,43){\scriptsize $\mathcal{L}_{\text{OCL}}$}
		\put(-222.5,36){\tiny (Eq.$_{\!}$~\ref{eq:11})}
		\put(-223,63){\scriptsize $\mathcal{L}_{\text{PCL}}$}
		\put(-221,56){\tiny (Eq.$_{\!}$~\ref{eq:7})}
		\end{center}
	\vspace{-18pt}
	\captionsetup{font=small}
	\caption{\small Diagram of our proposed space-time correspondence-aware training framework for matching-based VOS.
	}
	\label{fig:3}
	\vspace{-10pt}
\end{figure*}
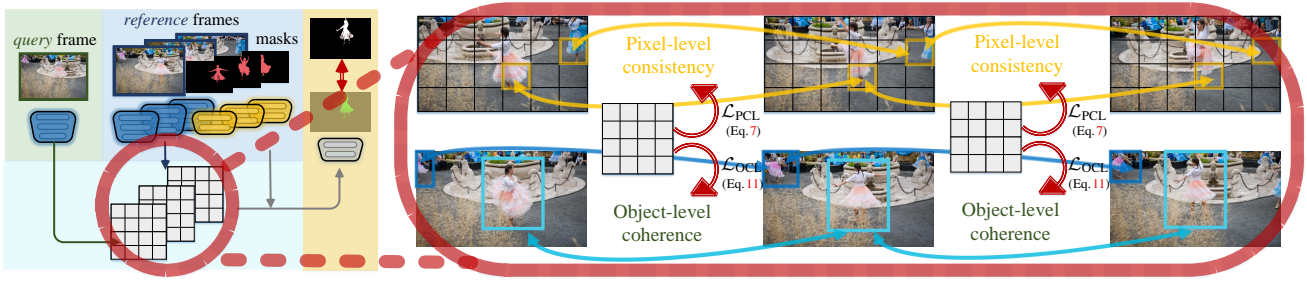

Next, a supportive feature $\bm{V}^{q\!}\!\in_{\!}\!\mathbb{R}^{D\!\times\! HW\!}$ for the query~can be created by aggregating memory value features using $A$:
\vspace{-2pt}
\begin{equation}\small
    \begin{aligned}\label{eq:4}
\bm{V}^{q}=\bm{V}^{r}\!A.
    \end{aligned}
    \vspace{-4pt}
\end{equation}
$\bm{V}^{q\!}$ is$_{\!}$ fed$_{\!}$ into$_{\!}$ a$_{\!}$ \textit{decoder}$_{\!}$ to$_{\!}$ output$_{\!}$ the$_{\!}$ mask$_{\!}$ $\hat{Y}^q$.$_{\!}$ $(I^q, \hat{Y}^q)_{\!}$~can be further added into the memory as new reference, and the query key is reused as the memory key. As the memory and decoder are not our focus, we refer to \cite{cheng2021rethinking,cheng2022xmem} for details.

\subsection{Space-time Correspondence-aware Training}\label{sec:3.2}
\noindent\textbf{Core Idea.} From Eq.$_{\!}$~\ref{eq:3} we can find that, the affinity $A$ gives the strength of all$_{\!}$ the$_{\!}$ pixel$_{\!}$ pairwise$_{\!}$ correlations$_{\!}$ between the query frame $I_{q}$ and$_{\!}$ the memorized reference frames ${I}_{r}$. Thus Eq.$_{\!}$~\ref{eq:3} essentially performs correspondence matching between the query and the memory (despite the normaliza- tion over all the reference frames), and the computed affini- ty$_{\!\!}$ $A_{\!}$ serves$_{\!}$ as$_{\!}$ the$_{\!}$ basis$_{\!}$ for$_{\!}$ the$_{\!}$ final$_{\!}$ mask$_{\!}$ decoding$_{\!}$~in$_{\!}$ Eq.$_{\!}$~\ref{eq:4}. Nevertheless, most existing matching-based VOS models are simply trained by minimizing the standard, supervised segmentation loss $\mathcal{L}_{\text{SEG}}$ (\textit{cf}.$_{\!}$~Eq.\ref{eq:2}). As a result, during training, the correspondence matching component (\textit{cf}.$_{\!}$~Eq.$_{\!}$~\ref{eq:2}) can only access the \textit{implicit}, segmentation-oriented supervision signals, yet lacking \textit{explicit} constraint/regulizarition over the cross-frame correlation estimation -- $A$. This may result in unreliable pixel association, hence suffering from sub-optimal performance eventually.


{Noticing the crucial role of space-time correspondence and the deficiency of standard training strategy in the context of matching-based VOS}, we hence {seek to complement} the$_{\!}$ segmentation-specified$_{\!}$ learning$_{\!}$ objective$_{\!}$ $\mathcal{L}_{\text{SEG}}$ (\textit{cf}.$_{\!}$~Eq.\ref{eq:2})$_{\!}$ with$_{\!}$ certain$_{\!}$ correspondence-aware$_{\!}$ training$_{\!}$ target.$_{\!}$ However,$_{\!}$ obtaining annotations of space-time correspondence for real videos$_{\!}$ is$_{\!}$ almost$_{\!}$ prohibitive,$_{\!}$ due$_{\!}$ to$_{\!}$ occlusions$_{\!}$ and$_{\!}$ free-form object$_{\!}$ deformations.$_{\!}$ This$_{\!}$ further$_{\!}$ motivates$_{\!}$ us$_{\!}$ to$_{\!}$ explore$_{\!}$~the intrinsic$_{\!}$ coherence$_{\!}$ of$_{\!}$ videos$_{\!}$ as$_{\!}$ a$_{\!}$ source$_{\!}$ of$_{\!}$ free$_{\!}$ supervision for$_{\!}$ correspondence$_{\!}$ matching.$_{\!}$ The$_{\!}$ delivered$_{\!}$ {outcome}$_{\!}$~is$_{\!}$~a powerful$_{\!}$ training$_{\!}$ framework$_{\!}$ that$_{\!}$ reinforces$_{\!}$ matching-based VOS models with \textit{annotation-free} correspondence learning.



Basically speaking, we holistically explore the coherence$_{\!}$ nature of video sequences on \textit{pixel} and \textit{object} granularities, within a contrastive correspondence learning scheme. 

\begin{figure}[t]
	\begin{center}
		\includegraphics[width=1.\linewidth]{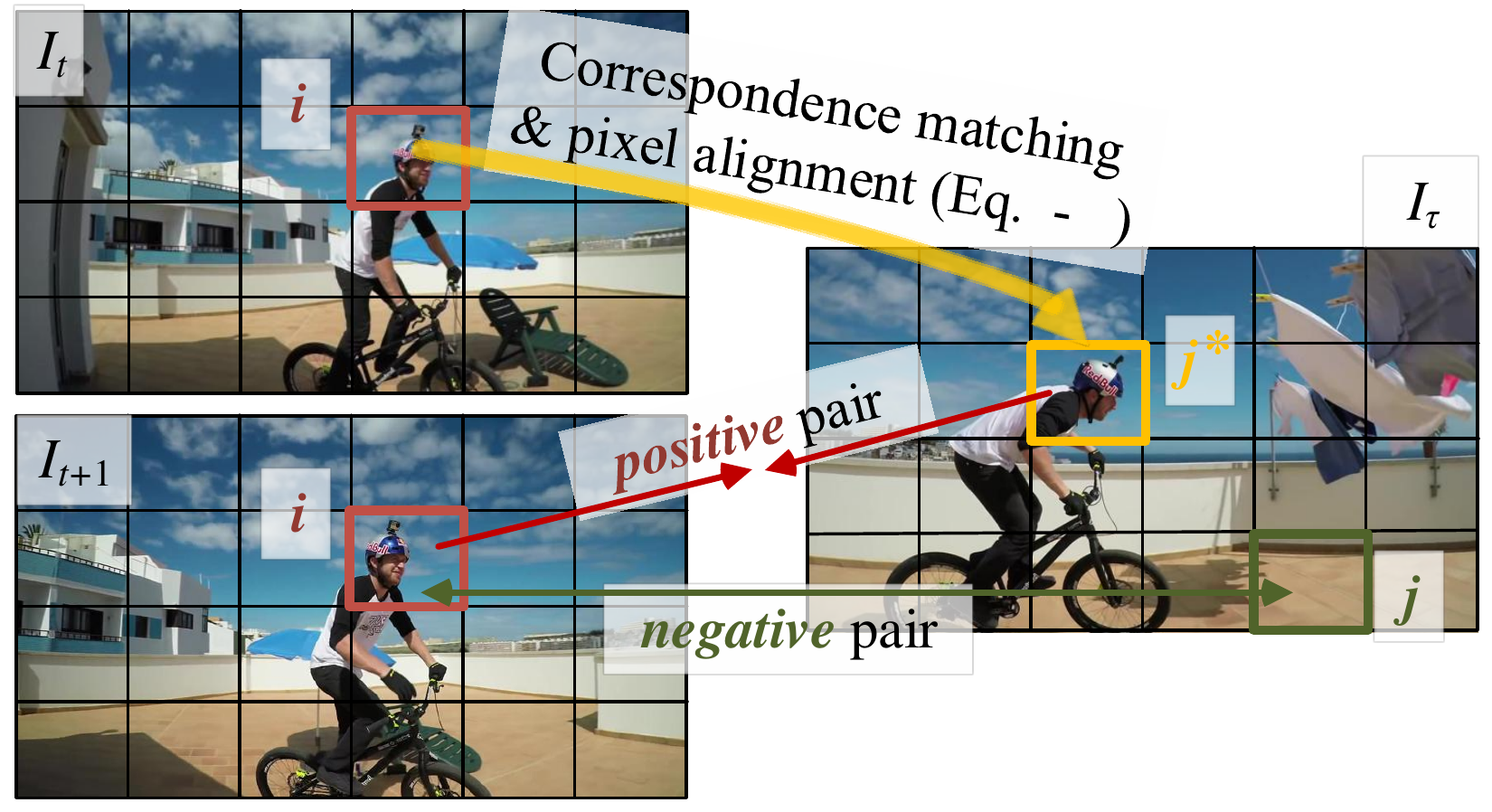}
		\put(-75,93.5){\small \rotatebox{-10}{\ref{eq:5}}}
		\put(-67,92.0){\small \rotatebox{-10}{\ref{eq:6}}}
		\end{center}
	\vspace{-18pt}
	\captionsetup{font=small}
	\caption{\small $_{\!}$Given$_{\!}$ two$_{\!}$ successive$_{\!}$ frames$_{\!}$ $I_{t}, I_{t+1\!}$ and$_{\!}$ an$_{\!}$ \textit{anchor}$_{\!}$ frame$_{\!}$ $I_{\tau}$, sampled from a same training video $\mathcal{I}$, we first make pixel-level correspondence matching between $I_{t}$ and $I_{\tau}$, through Eq.~\ref{eq:5}-\ref{eq:6}. Then the alignment results are used as the  pseudo label for the contrastive correspondence learning (\ie, Eq.$_{\!}$~\ref{eq:7}) between $I_{t+1}$ and $I_{\tau}$, based on the local continuity assumption, \ie, $I_{t}$ and $I_{t+1}$ yield consistent patterns at spatially adjacent locations. For clarity, negative pixel/patch samples from other training videos are omitted.
	}
	\label{fig:4}
	\vspace{-10pt}
\end{figure}

\noindent \textbf{Correspondence learning based on Pixel-level Consis- tency.$_{\!}$} We$_{\!}$ first$_{\!}$ address$_{\!}$ local$_{\!}$ continuity$_{\!}$ {residing}$_{\!}$ in$_{\!}$ videos,$_{\!}$ \ie, spatiotemporally$_{\!}$ adjacent$_{\!}$ pixels/patches$_{\!}$ typically$_{\!}$ yield$_{\!}$ consistent$_{\!}$ patterns. Specifically, given a training video $\mathcal{I}$, we sample two successive frames $I_{t}, I_{t+1}$ as well as an \textit{anchor} frame $I_{\tau}$, where $\tau_{\!}\neq_{\!}t$ and  $\tau_{\!}\neq_{\!}t_{\!}+_{\!}1$. Based on a contrastive formulation, our approach estimates pixel pairwise correlations of $I_{t\!}$ and $I_{t+1\!}$ \textit{w.r.t.}$_{\!}$~the anchor $I_{\tau}$, and learns correspondence$_{\!}$ matching$_{\!}$~by$_{\!}$ enforcing$_{\!}$ these$_{\!}$ correlations$_{\!}$ to$_{\!}$ be$_{\!}$ spatially consistent across $I_{t}$ and $I_{t+1}$.$_{\!}$ More precisely, with the key$_{\!}$ encoder$_{\!}$ $\kappa$,$_{\!}$ we$_{\!}$ have$_{\!}$ the$_{\!}$ dense$_{\!}$ visual$_{\!}$ representations,$_{\!}$~\ie,
$\bm{K}_{t}, \bm{K}_{t+1}, \bm{K}_{\tau}\!\in\!\mathbb{R}^{C\!\times\! HW\!}$, of $I_{t}, I_{t+1}$, and $I_{\tau}$, respectively. Then, we compute the affinity of the pixel/patch feature vector at $i$-th position of $\bm{K}_{t}$ \textit{w.r.t.}$_{\!}$~the anchor feature tensor $\bm{K}_{\tau}$:
\vspace{-5pt}
\begin{equation}\small
    \begin{aligned}\label{eq:5}
A^{t, \tau}(i, j)\!=\!\frac{\exp\big(\langle\bm{K}_{t}(i), \bm{K}_{\tau}(j)\rangle\big)}{\textstyle\sum\nolimits_{j'}\exp\big(\langle\bm{K}_{t}({i}), \bm{K}_{\tau}(j')\rangle\big)}. 
    \end{aligned}
    \vspace{-0pt}
\end{equation}
We acquire the pixel/patch $j^*$ in $\bm{K}_{\tau}$ that best matches the pixel/patch $i$ in $\bm{K}_{t}$ (see Fig.~\ref{fig:4}):
\vspace{-3pt}
\begin{equation}\small
    \begin{aligned}\label{eq:6}
j^*=\underset{j \in \{1, \cdots_{\!}, HW\}}{\arg \max } A^{t, \tau}(i, j).
    \end{aligned}
    \vspace{-3pt}
\end{equation}
The index of the best alignment $j^*$ of $\bm{K}_{t}(i)$ is subtly used as the pseudo label for correspondence matching between the $i$-th  feature vector of $\bm{K}_{t+1}$ and the anchor feature tensor $\bm{K}_{\tau}$, hence addressing local consistency and enabling self-supervised$_{\!}$ learning$_{\!}$ of$_{\!}$ pixel-level$_{\!}$ correspondence$_{\!}$ matching:
\vspace{-4pt}
\begin{equation}\small
    \begin{aligned}\label{eq:7}
\mathcal{L}_{\text{PCL}}\!=-\log\!\sum\nolimits_i\frac{\exp_{\!}\big(\langle\bm{K}_{t+1}(i), \bm{K}_{\tau}(j^*)\rangle\big)}{\textstyle\sum\nolimits_{j}\exp_{\!}\big(\langle\bm{K}_{t+1}({i}),  \bm{K}_{\tau}(j)\rangle\big)}. 
    \end{aligned}
\end{equation}
Such self-supervised loss trains the model to distinguish~the aligned$_{\!}$ pair,$_{\!}$ \ie,$_{\!}$ $(\bm{K}_{t+1}(i),_{\!} \bm{K}_{\tau}(j^*))$,$_{\!}$ from$_{\!}$ the$_{\!}$ set$_{\!}$ of non-corresponding$_{\!}$ ones,$_{\!}$ \ie,$_{\!}$  $\{(\bm{K}_{t+1}({i}),_{\!} \bm{K}_{\tau}(j))\}_{j\neq j^*}$,$_{\!}$  based$_{\!}$  on the$_{\!}$ assignment$_{\!}$ of$_{\!}$ $\bm{K}_{t}(i)$, which is located at the same spatial position of$_{\!}$ $\bm{K}_{t+1}(i)$,$_{\!}$ to$_{\!}$ the$_{\!}$ anchor$_{\!}$ $\bm{K}_{\tau}$.$_{\!}$ Through$_{\!}$ this$_{\!}$~self- training$_{\!}$ mechanism,$_{\!}$ the$_{\!}$ model$_{\!}$ learns$_{\!}$ to$_{\!}$ assign$_{\!}$ the$_{\!}$ features in$_{\!}$~frame$_{\!}$ $I_{t\!}$ consistently$_{\!}$ with$_{\!}$ the$_{\!}$ temporally$_{\!}$ proximate$_{\!}$ frame $I_{t+1}$ \textit{w.r.t.}$_{\!}$~$I_{\tau}$, so as to impose the desired property of local consistency on the visual embedding space $\kappa$ and encourage reliable correspondence matching explicitly. Moreover, following the common practice of contrastive learning \cite{chen2020simple,he2020momentum}, we$_{\!}$ randomly$_{\!}$ sample$_{\!}$ pixel/patch$_{\!}$ features$_{\!}$ from$_{\!}$ other$_{\!}$ videos$_{\!}$ in~the training batch as negative examples during the computation  of $\mathcal{L}_{\text{PCL}}$, teaching the model to efficiently disambi- guate correspondence on both inter- and intra-video levels.


\noindent \textbf{Correspondence Learning based on Object-level Cohe- rence.} Through implementing the pixel-level consistency$_{\!}$ property$_{\!}$ in$_{\!}$ our$_{\!}$ framework,$_{\!}$ we$_{\!}$ inspire$_{\!}$ matching-based$_{\!}$~VOS models$_{\!}$ to$_{\!}$ learn$_{\!}$ locally$_{\!}$ distinctive$_{\!}$ features, hence constructing  reliable,$_{\!}$ dense$_{\!}$ correspondence$_{\!}$ between$_{\!}$ video$_{\!}$ frames. {For the sake of full-scale robust matching},$_{\!}$ we$_{\!}$ further$_{\!}$ investi- gate$_{\!}$ the$_{\!}$ content$_{\!}$ continuity of videos on the object-level -- representations of a same object instance should remain stable across frames. By enforcing the key encoder {$\kappa$} to learn object-level$_{\!}$ compact$_{\!}$ and$_{\!}$ discriminative$_{\!}$ representations, we are able to boost the robustness of correspondence matching against local disturbance caused by deformation and occlusion, and better address the object-aware nature of the VOS task.$_{\!}$ Put$_{\!}$ simply,$_{\!}$ we$_{\!}$ apply$_{\!}$ contrastive$_{\!}$ correspondence$_{\!}$~learn-
ing$_{\!}$ on$_{\!}$ both$_{\!}$ automatically$_{\!}$ discovered$_{\!}$ and$_{\!}$ pre-labeled$_{\!}$ video objects; object-level, space-time correspondence$_{\!}$ is$_{\!}$ learnt$_{\!}$ by$_{\!}$  maximizing$_{\!}$ the$_{\!}$ similarity$_{\!}$ of$_{\!}$ the$_{\!}$ representations$_{\!}$ of$_{\!}$ the$_{\!}$ same$_{\!}$ object instance at different timesteps.

VOS training videos often involve complex visual scenes with multiple objects, while only a small portion of the object instances are labeled$_{\!}$~\cite{pont2017davis,xu2018youtube} (\ie, the masked objects in Fig.$_{\!}$~\ref{fig:4} (a-b)).$_{\!}$ We$_{\!}$ thus adopt Selective Search$_{\!}$~\cite{uijlings2013selective}, an unsupervised, object proposal algorithm, to obtain an exhaustive set of potential {objects} for each training frame $I$ (plotted~in solid boxes in Fig.$_{\!}$~\ref{fig:5} (a-b)).$_{\!}$ The$_{\!}$ automatically$_{\!}$ discovered object proposals substantially provide diverse training$_{\!}$ samples to$_{\!}$ aid$_{\!}$ object-level$_{\!}$ correspondence$_{\!}$ learning,$_{\!}$ without$_{\!}$ adding\\
\noindent extra$_{\!}$ annotation$_{\!}$ cost.$_{\!}$ Formally,$_{\!}$ let$_{\!}$ $\mathcal{P}_{\!}\!=_{\!}\!\{P_i\}_{i\!}$ {denote}$_{\!}$ the$_{\!}$ full set of automatically discovered objects and pre-labeled ones in frame $I$. Each object $P\!=\!(x, y, w, h)$ is represented as a bounding box, whose center is $(x, y)$ and the size is of $w\!\times\!h$, and its object-level representation $\bm{p}\!\in\!\mathbb{R}^D$ is given by:
\vspace{-4pt}
\begin{equation}\small
    \begin{aligned}
        \bm{p}=\texttt{AVGPool}(\texttt{RoIAlign}(\bm{K}, P)).
    \end{aligned}
    \vspace{-4pt}
    \label{eq:8}
\end{equation}

Given two distant frames, $I_t$ and $I_{t'}$, sampled from training video $\mathcal{I}$, as well as their corresponding object sets, \ie, $\mathcal{P}$ and $\mathcal{P}'$, a small set of objects are first drawn from  $\mathcal{P}$, \ie, $\mathcal{Q}\!\subset\!\mathcal{P}$; the objects are sampled as the spatial distribution of the centers are sparse, so as to ensure the objects are from different instances. Note that we make  $|\mathcal{P}'|\!\gg\!|\mathcal{Q}|$ to ensure $\mathcal{P}'$ covers all the instances appeared in $\mathcal{Q}$. Given $\mathcal{Q}$, we are next to find the corresponding objects in $\mathcal{P}'$, which can be formulated as a bipartite matching problem\footnote{We only make bipartite matching for the automatically discovered~ob- jects,$_{\!}$ as the$_{\!}$ correspondence$_{\!}$ between$_{\!}$ the$_{\!}$ annotated ones is already known.} (see Fig.~\ref{fig:5} (c)):
\vspace{-4pt}
\begin{equation}\small
    \begin{aligned}
		& \max_{A^{t,t'}} \sum\nolimits_{i} \sum\nolimits_{j\!}\langle\bm{p}_{i}, \bm{p}'_{j}\rangle \cdot A^{t,t'\!}(i,j), \\
		\!\!\!\!\!\textit{s.t.}~~~~ & \forall{p_{i}\!\in\!\mathcal{Q}},~~~~\sum\nolimits_{j=1}A^{t,t'\!}(i,j) \leq 1, \\
		& \forall{p'_{j}\!\in\!\mathcal{P}'\!},~~~~\sum\nolimits_{i=1\!}A^{t,t'\!}(i,j) \leq 1,\\
		&\forall(p_{i}, p'_{j}), ~~~~A^{t,t'\!}(i,j)\in\{0,1\},
    \end{aligned}
    \vspace{-2pt}
    \label{eq:9}
\end{equation}
where $A^{t,t'\!}\!\in\!\{0,1\}^{|\mathcal{Q}|\times|\mathcal{P}'|\!}$ refers to the assignment of $\mathcal{Q}$ \textit{w.r.t.}\!~$\mathcal{P}'$, and the constraints ensure exclusive assignment. The global optimal solution of Eq.\!~\ref{eq:9} can be acquired by the Hungarian algorithm\!~\cite{kuhn1955hungarian}, and is cleverly leveraged as the pseudo label for our object-level correspondence learning.

\begin{figure}[t]
	\begin{center}
		\includegraphics[width=1.\linewidth]{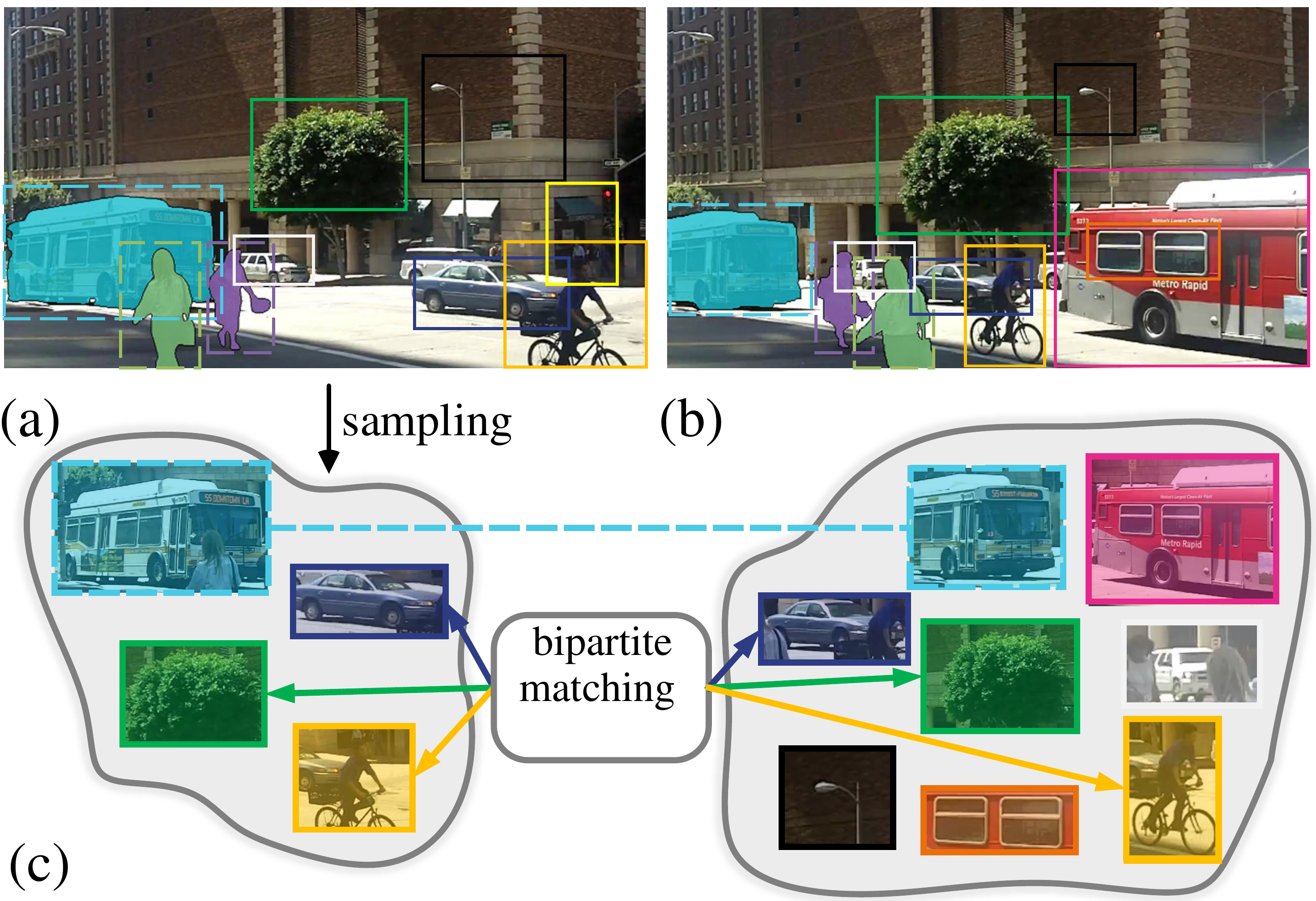}
              \put(-137,28){\scriptsize (Eq.$_{\!}$~\ref{eq:9})}
       \put(-14,7){\small $\mathcal{P}'$}
       \put(-217,13){\small $\mathcal{Q}$}
		\end{center}
	\vspace{-18pt}
	\captionsetup{font=small}
	\caption{\small (a-b) Frames $I_t$ and $I_{t'}$ with their corresponding object sets, \ie, $\mathcal{P}$ and $\mathcal{P}'$, where the manually annotated object instances are$_{\!}$ plotted$_{\!}$ in$_{\!}$ dashed$_{\!}$ boxes$_{\!}$ and$_{\!}$ the$_{\!}$ object$_{\!}$ proposals$_{\!}$ discovered$_{\!}$~by Selective Search$_{\!}$~\cite{uijlings2013selective} are plotted in solid boxes. (c) Bipartite mat- ching$_{\!}$ (Eq.$_{\!}$~\ref{eq:9})$_{\!}$ is$_{\!}$ made$_{\!}$ between$_{\!}$ a$_{\!}$ subset$_{\!}$ of$_{\!}$ $\mathcal{P}$,$_{\!}$ \ie,$_{\!}$ $\mathcal{Q}\!\subset\!\mathcal{P}$,$_{\!}$ and$_{\!}$ the full$_{\!}$ object$_{\!}$ set$_{\!}$ $\mathcal{P}'$,$_{\!}$ so$_{\!}$ as$_{\!}$ to$_{\!}$ find$_{\!}$ paired$_{\!}$ objects$_{\!}$ in$_{\!}$  $I_{t\!}$ and$_{\!}$ $I_{t'}$.$_{\!}$ The$_{\!}$ object pairs are used as positive samples for our object-level coherence based contrastive correspondence learning  (\ie, Eq.$_{\!}$~\ref{eq:11}).  For clarity, negative object samples from other training videos are omitted.
	}
	\label{fig:5}
	\vspace{-12pt}
\end{figure}

$_{\!}$Concretely,$_{\!}$ for$_{\!}$ each$_{\!}$ object$_{\!}$ $p_{i}\!\in_{\!}\!\mathcal{Q}$,$_{\!}$ the$_{\!}$ index$_{\!}$ of$_{\!}$ its$_{\!}$ aligned counterpart  in $\mathcal{P}'\!$ can be directly derived from  $A^{t,t'\!}$:
\vspace{-3pt}
\begin{equation}\small
    \begin{aligned}\label{eq:10}
j^*=\underset{j \in \{1, \cdots_{\!}, |\mathcal{P}'|\}}{\arg \max } A^{t,t'\!}(i,j).
    \end{aligned}
    \vspace{-3pt}
\end{equation}
Given a query object $p_{i}\!\in\!\mathcal{Q}$, we view $p'_{j^*}\!\in\!\mathcal{P}'$ as a positive sample while objects from other videos in the training batch as negative samples (denoted as $\mathcal{O}\!=\!\{O_1, O_2, \cdots\}$). Hence the contrastive learning objective of our object-level correspondence can be formulated as:
\vspace{-4pt}
\begin{equation}\small
    \begin{aligned}\label{eq:11}
		\hspace{-2mm}
\mathcal{L}_{\text{OCL}}\!=-\log\!\sum_{p_{i}\in\mathcal{Q}}\frac{\exp_{\!}\big(\langle\bm{p}_i, \bm{p}'_{j^*}\rangle\big)}{\exp_{\!}\big(\langle\bm{p}_i, \bm{p}'_{j^*}\rangle\big)\!+\!\textstyle\sum\nolimits_{o\in\mathcal{O}}\exp_{\!}\big(\langle\bm{p}_{i},  \bm{o}\rangle\big)}. 
    \end{aligned}
\end{equation}
By contrasting the positive object pair, \ie, $({p}_{i}, {p}'_{j^*})$, against negative ones, \ie, $\{({p}_{i}, {o})\}_{o\in\mathcal{O}}$, features of same object instances in different video frames are forced to be aligned. In$_{\!}$ this$_{\!}$ way,$_{\!}$ our$_{\!}$ training$_{\!}$ framework$_{\!}$ introduces$_{\!}$ the$_{\!}$ object-level consistency property into the visual embedding space $\kappa$ of matching-based$_{\!}$ segmentation$_{\!}$ models,$_{\!}$  facilitating$_{\!}$ the$_{\!}$ disco-\\
\noindent  very$_{\!}$ of$_{\!}$ robust,$_{\!}$ object-oriented$_{\!}$ correspondence.$_{\!}$ In$_{\!}$ practice, we find that our pixel-level and object-level correspondence learning strategies (\ie, Eq.$_{\!}$~\ref{eq:7} and Eq.$_{\!}$~\ref{eq:11}) boost the performance \textit{collaboratively} (see related experiments in \S\ref{sec:ede}).

\subsection{Implementation Details}\label{sec:3.3}



\noindent \textbf{Network$_{\!}$ Configuration.$_{\!\!}$} We$_{\!}$ apply$_{\!}$ our$_{\!}$ training$_{\!}$ algorithm to~two$_{\!}$ top-leading$_{\!}$ matching-based$_{\!}$ VOS$_{\!}$ models:$_{\!}$ STCN$_{\!\!}$~\cite{cheng2021rethinking} and$_{\!}$ XMem$_{\!}$~\cite{cheng2022xmem},$_{\!}$ without$_{\!}$ architecture$_{\!}$ change.$_{\!}$ Specifically,$_{\!}$~as in$_{\!}$~\cite{cheng2021rethinking,cheng2022xmem} the key encoder $\kappa$ and value encoder $\upsilon$ are constructed with the first four residual blocks of ResNet50$_{\!}$~\cite{he2016deep} and ResNet18, respectively. A $3\!\times\!3$ convolutional layer~is used$_{\!}$ to$_{\!}$ project$_{\!}$ the$_{\!}$ \texttt{res4}$_{\!}$ feature$_{\!}$ with$_{\!}$ stride$_{\!}$ 16$_{\!}$ to$_{\!}$ either$_{\!}$ the key$_{\!}$ feature$_{\!\!}$ $\bm{K}_{\!}$ of$_{\!}$ $C\!=\!64_{\!}$ channels$_{\!}$ or$_{\!}$ the$_{\!}$ value$_{\!\!}$ $\bm{V}_{\!}$ of$_{\!}$ $D\!=\!512$ channels.$_{\!}$ For$_{\!}$ the$_{\!}$ decoder,$_{\!}$ it$_{\!}$ first$_{\!}$ compresses$_{\!}$ the$_{\!}$ memory$_{\!}$ out-$_{\!}$ put$_{\!}$ $\bm{V}^{q\!}$ (\textit{cf}.$_{\!}$~Eq.$_{\!}$~\ref{eq:4}) to 512 channels with a residual block, and then makes gradual upsampling by $2\times$ until stride 4  with higher-resolution features from $\kappa$ incorporated using skip- connections.$_{\!}$ The$_{\!}$ memory$_{\!}$ stores$_{\!}$ features$_{\!}$ of$_{\!}$ past$_{\!}$ frames$_{\!}$ for$_{\!}$ long-term$_{\!}$ modeling; details can be referred to$_{\!}$~\cite{cheng2021rethinking,cheng2022xmem}.

\noindent \textbf{Pixel-level Correspondence Learning.} For training efficiency and robustness, we consider a sparse set of features of the anchor frame $I_{\tau}$, which are obtained through random sampling over $I_{\tau\!}$ with a spatially uniform grid of size $8\!\times\!8$,\\
\noindent instead of using all the point features of $I_{\tau}$, during the computation of Eq.$_{\!}$~\ref{eq:5}. {The sampled pixels from the same batch are utilized for contrastive learning (the$_{\!}$ ratio$_{\!}$ of positive and negative pair is 1/$5{\!}\times{\!}10^4$).} In addition to sampling $I_{\tau}$ from $\mathcal{I}$, we apply random multi-scale crop and horizontal flipping to $I_{t}$ and treat the transformed frame as an anchor frame. This allows$_{\!}$ us to$_{\!}$ approach$_{\!}$ cross-frame$_{\!}$ and$_{\!}$ cross-view$_{\!}$ contrastive correspondence learning within a unified framework.

\noindent  \textbf{Object-level Correspondence Learning.} Considering the redundancy of the proposals generated by Selective Search \cite{uijlings2013selective} (typically thousands of proposals per frame image), we follow the common practice in object-level self-supervised representation learning$_{\!}$~\cite{wei2021aligning,xie2021unsupervised} to keep only the proposals $P\!=\!(x, y, w, h)$  that satisfy: i) the aspect ratio $w/h$ is between 1/3 and 3/1; and ii) the scale $w\!\times\!h$
occupies between $0.3^{2\!}$ and$_{\!}$ $0.8^{2\!}$ of$_{\!}$ the$_{\!}$ entire$_{\!}$ image$_{\!}$ area.$_{\!}$ Moreover,$_{\!}$ we$_{\!}$ split$_{\!}$ the lattice of the frame image into multiple $32\!\times\!32$ grids, and cluster proposals into the grids which the center positions $(x, y)$ fall in. The proposals in $\mathcal{Q}$ are sampled from the~ob- ject$_{\!}$ clusters$_{\!}$ in$_{\!}$ $I_{t\!}$ (up$_{\!}$ to$_{\!}$ sampling$_{\!}$ one$_{\!}$ proposal$_{\!}$ for$_{\!}$ each$_{\!}$ clu-$_{\!}$ ster),$_{\!}$ to$_{\!}$ omit$_{\!}$ duplicated$_{\!}$ proposals.$_{\!}$ We$_{\!}$ empirically$_{\!}$ set$_{\!}$ $|\mathcal{Q}|\!=\!3$.$_{\!\!}$

\noindent \textbf{Training$_{\!}$ Objective.$_{\!}$} The$_{\!}$ final$_{\!}$ learning$_{\!}$ target$_{\!}$ is$_{\!}$ the$_{\!}$ combina-$_{\!}$  tion$_{\!}$ of$_{\!}$ the$_{\!}$ standard$_{\!}$ VOS$_{\!}$ training$_{\!}$ objective$_{\!}$ $\mathcal{L}_\text{SEG\!}$ (\textit{cf.}$_{\!}$~Eq.$_{\!}$~\ref{eq:2}) and$_{\!}$ our$_{\!}$ proposed$_{\!}$ two$_{\!}$ self-supervised$_{\!}$ correspondence$_{\!}$ learn- ing$_{\!}$ loss$_{\!}$ functions,$_{\!}$~\ie,
 $\mathcal{L}_\text{PCL\!}$ (pixel-level;$_{\!}$ Eq.$_{\!}$~\ref{eq:7})$_{\!}$ and$_{\!}$  $\mathcal{L}_\text{OCL\!}$ (object-level;$_{\!}$ Eq.$_{\!}$~\ref{eq:11}):
\vspace{-3pt}
\begin{equation}\small
    \begin{aligned}\label{eq:12}
    \mathcal{L} = \mathcal{L}_\text{SEG} + \alpha(\mathcal{L}_\text{PCL} + \beta\mathcal{L}_\text{OCL}),
    \end{aligned}
    \vspace{-4pt}
\end{equation}
where the coefficient $\alpha\!\in\![0,0.2]$ is scheduled following a linear warmup policy and $\beta$ is fixed as 0.5.

\section{Experiment}

\begin{table}
	\centering
	\small
	\resizebox{\columnwidth}{!}{
		\setlength\tabcolsep{4pt}
		\renewcommand\arraystretch{1.1}
		\begin{tabular}{lcccc||ccc}
			\hline\thickhline
		    \rowcolor{mygray}
		     & & \multicolumn{3}{c||}{\text{DAVIS2017}$_{val}$} & \multicolumn{3}{c}{\text{DAVIS2017}$_{test-dev}$} \\ \cline{3-8}
            \rowcolor{mygray}
            \multirow{-2}{*}{Method} & \multirow{-2}{*}{S} & $\mathcal{J}$\&$\mathcal{F}_m$ $\uparrow$ & $\mathcal{J}$ $\uparrow$ & $\mathcal{F}$ $\uparrow$ & $\mathcal{J}$\&$\mathcal{F}_m$ $\uparrow$ & $\mathcal{J}$ $\uparrow$ & $\mathcal{F}$ $\uparrow$ \\  \hline \hline

			FEELVOS\!~\cite{voigtlaender2019feelvos}  & \xmark & 71.6  & 69.1 & 74.0  & 57.8 & 55.2  & 60.5 \\
			SSTVOS\!~\cite{duke2021sstvos} & \xmark & 82.5 & 79.9 & 85.1  & - & - & - \\
			CFBI+\!~\cite{yang2021collaborative}  & \xmark & 82.9 & 80.1 & 85.7 & 75.6 & 71.6 & 79.6 \\
			Joint\!~\cite{mao2021joint}  & \xmark  & 83.5 & 80.8 & 86.2   & - & - & - \\\hline
			STCN \!~\cite{cheng2021rethinking}  &  & 82.5 & 79.3 & 85.7 & 73.9 & 69.9 & 77.9 \\
			STCN+\textbf{\texttt{Ours}}  & \multirow{-2}{*}{\xmark}  & \textbf{84.7} &	\textbf{81.6} & \textbf{87.8} &  \textbf{77.3}	& \textbf{73.5}	& \textbf{81.1}   \\  \cdashline{1-8}[1pt/1pt]
			XMem\!~\cite{cheng2022xmem}  &   & 84.5 & - & - & 79.8 & - &  -  \\
			XMem+\textbf{\texttt{Ours}}  & \multirow{-2}{*}{\xmark}  &  \textbf{86.1} &  \textbf{82.7}	& \textbf{89.5}	& \textbf{81.0} & \textbf{77.3}	 & \textbf{84.7}  \\
			\hline
            \hline
			STM\!~\cite{oh2019video}  & \cmark  & 81.8 & 79.2 & 84.3 & 72.2 & 69.3 & 75.2  \\
			EGMN\!~\cite{lu2020video}  & \cmark  & 82.8 & 80.2 & 85.2  & - & - & -  \\
			KMN\!~\cite{seong2020kernelized}  & \cmark  & 82.8 & 80.0 & 85.6  & 77.2 & 74.1 & 80.3 \\
			RMNet\!~\cite{xie2021efficient}  & \cmark  & 83.5 & 81.0 & 86.0  & 75.0 & 71.9 & 78.1 \\
			LCM\!~\cite{hu2021learning}  & \cmark  & 83.5 & 80.5 & 86.5  & 78.1 & 74.4 & 81.8 \\
			HMMN\!~\cite{seong2021hierarchical}  & \cmark & 84.7 & 81.9 & 87.5 & 78.6 & 74.7 &82.5 \\
			AOT\!~\cite{yang2021associating}  & \cmark  & 84.9 & 82.3 & 87.5  & 79.6 & 75.9 & 83.3 \\
			RDE\!~\cite{li2022recurrent}  & \cmark  & 84.2 & 80.8 & 87.5 & 77.4 & 73.6 & 81.2 \\
			PCVOS\!~\cite{park2022per} & \cmark  & 86.1 & 83.0 & 89.2 & - & - & -  \\
			{DeAOT\!~\cite{yang2022decoupling}}  & \cmark  & {86.2} & {83.1} &{89.3} & {77.5} & {74.0}  & {80.9}   \\
			\hline
			STCN\!~\cite{cheng2021rethinking}  &  & 85.4 & 82.2 & 88.6 & 76.1 & 73.1 & 80.0  \\
			STCN+\textbf{\texttt{Ours}}  & \multirow{-2}{*}{\cmark}  & \textbf{86.6}  & \textbf{83.0} & \textbf{90.1}	&  \textbf{79.3}  &  \textbf{75.8}  &  \textbf{82.8} \\\cdashline{1-8}[1pt/1pt]	
			XMem\!~\cite{cheng2022xmem}  &  & 86.2 & 82.9 & 89.5  & 81.0 & 77.4 & 84.5 \\
			XMem+\textbf{\texttt{Ours}}  & \multirow{-2}{*}{\cmark}  &  \textbf{87.7}  & \textbf{84.1} & \textbf{91.2}	&  \textbf{82.0}	& \textbf{78.3}	 &  \textbf{85.6}  \\
			\hline
		\end{tabular}
	}
	\captionsetup{font=small}
	\caption{\small{Results$_{\!}$ on$_{\!}$ DAVIS2017$_{val\!}$ and$_{\!}$ DAVIS2017$_{test-dev\!}$~\cite{pont2017davis}$_{\!}$ (\S\ref{sec:ecs}). S: if synthetic data is used for pre-training.}}
	\label{table:davis}
	\vspace{-11pt}
\end{table}

\subsection{Experimental Setup}

\noindent\textbf{Datasets.} We give extensive experiments on three datasets.
\begin{itemize}[leftmargin=*]
	\setlength{\itemsep}{0pt}
	\setlength{\parsep}{-2pt}
	\setlength{\parskip}{-0pt}
	\setlength{\leftmargin}{-10pt}
	\vspace{-6pt}
	\item \textbf{DAVIS2016}$_{\!}$~\cite{perazzi2016benchmark}$_{\!}$ has$_{\!}$ $50_{\!}$ single-object$_{\!}$ videos$_{\!}$ that$_{\!}$ are$_{\!}$~fine- ly labeled at $24$ FPS and split into $30/20$ for \texttt{train}/\texttt{val}.$_{\!}$

	\item \textbf{DAVIS2017}$_{\!}$~\cite{pont2017davis} contains  $60/30$ multi-object videos for $\texttt{train}/\texttt{val}$. It also provides a \texttt{test-dev} set consisting of $30$ videos with more challenging scenarios.
	
	\item \textbf{YouTube-VOS}$_{\!}$~\cite{xu2018youtube} includes $3,471$ videos for training and $474/507$ videos for validation in the $2018/2019$ split, respectively. The videos are sampled at $30$ FPS and annotated per $5$ frame with single or multiple objects.

	
	\vspace{-6pt}
\end{itemize}

\noindent \textbf{Training.} For fair comparison, we adopt the standard training protocol \cite{oh2019video,lu2020video,cheng2021rethinking,yang2021associating}, which has two phases: \textbf{First}, we pre-train the network on synthetic videos generated from static, segmentation images$_{\!}$~\cite{wang2017learning, shi2015hierarchical, zeng2019towards, cheng2020cascadepsp, li2020fss}. \textbf{Second}, the main$_{\!}$ training$_{\!}$ is$_{\!}$ made$_{\!}$ on$_{\!}$ DAVIS2017$_{train\!}$ and$_{\!}$ YouTube-VOS2019$_{train}$. At each training step, we sample $3$ frames per {video} to create mini-sequences, as in$_{\!}$~\cite{cheng2021rethinking,cheng2022xmem}. More training details can be found in the supplementary.


\noindent\textbf{Testing.} 
All the configurations in the testing phase are kept exactly the same as the baseline. Note that our algorithm~is only applied at the training time; it renders no redundant computation load and speed delay to deployment process, equally efficient as the baseline models.

\noindent\textbf{Evaluation.} We follow the official evaluation protocol$_{\!}$~\cite{perazzi2016benchmark} to adopt region similarity$_{\!}$~($\mathcal{J}$), contour accuracy$_{\!}$~($\mathcal{F}$), and their$_{\!}$ average$_{\!}$ score$_{\!}$~($\mathcal{J}\&\mathcal{F}_m$)$_{\!}$ for$_{\!}$ evaluation.$_{\!}$ Performance$_{\!}$~on DAVIS2017$_{test\text{-}dev\!}$ and YouTube-VOS2018$_{val}$\!~\&\!~2019$_{val\!}$ is obtained by submitting the results to the official servers;~the\\ \noindent latter$_{\!}$ two$_{\!}$ sets$_{\!}$ are$_{\!}$ further$_{\!}$ reported$_{\!}$ at$_{\!}$ \textit{seen}$_{\!}$ and$_{\!}$ \textit{unseen}$_{\!}$ classes.


\subsection{Comparison to State-of-the-Arts}\label{sec:ecs}

\noindent\textbf{DAVIS2016$_{\!}$~\cite{perazzi2016benchmark}.$_{\!}$} As$_{\!}$ demonstrated$_{\!}$ in$_{\!}$ Table$_{\!}$~\ref{table:davis16},$_{\!}$ our$_{\!}$ approa- ch makes stable performance gains over STCN (91.6\% $\rightarrow$ \textbf{92.0\%}) and XMem (91.5\%$\rightarrow$\textbf{92.2\%}) on$_{\!}$ DAVIS2016$_{val}$, and outperforms all the previous state-of-the-arts. Such results are particularly impressive, considering DAVIS2016 is a simple yet extensively studied dataset.

\begin{table}[t]
    \small\center
	\resizebox{0.80\columnwidth}{!}{
		\setlength\tabcolsep{6pt}
		\renewcommand\arraystretch{1.1}
		\begin{tabular}{lcccc}
			\hline\thickhline
		    \rowcolor{mygray}
		     & & \multicolumn{3}{c}{\text{DAVIS2016}$_{val}$} \\ \cline{3-5}
            \rowcolor{mygray}
            \multirow{-2}{*}{Method} & \multirow{-2}{*}{Synthetic} & $\mathcal{J}$\&$\mathcal{F}_m$ $\uparrow$ & $\mathcal{J}$ $\uparrow$ & $\mathcal{F}$ $\uparrow$ \\  \hline \hline
			RMNet\!~\cite{xie2021efficient}  & \cmark  & 88.8 & 88.9 & 88.7  \\
			STM\!~\cite{oh2019video}  & \cmark  & 89.3 & 88.7 & 89.9  \\
			LCM\!~\cite{hu2021learning}  & \cmark  & 90.7 & 89.9 & 91.4  \\
			HMMN\!~\cite{seong2021hierarchical}  & \cmark & 90.9 & 89.6 & 92.0 \\
			AOT\!~\cite{yang2021associating}  & \cmark  & 91.1 & 90.1 & 92.1  \\
			RDE\!~\cite{li2022recurrent}  & \cmark  & 91.1 & 89.7 & 92.5  \\
			PCVOS\!~\cite{park2022per}     & \cmark & 91.9 & 90.8 & 93.0 \\
			 \hline
			STCN\!~\cite{cheng2021rethinking}  &  & 91.6 & 90.8 & 92.5  \\
			STCN+\textbf{\texttt{Ours}}	& \multirow{-2}{*}{\cmark} & \textbf{92.0}  & \textbf{91.0} & \textbf{92.9} \\	\cdashline{1-5}[1pt/1pt]	
			XMem\!~\cite{cheng2022xmem}  &   & 91.5 & 90.4 & 92.7 	\\
			XMem+\textbf{\texttt{Ours}}  & \multirow{-2}{*}{\cmark}  & \textbf{92.2} & \textbf{91.1} & \textbf{93.3}  \\
			\hline
		\end{tabular}
		
	}
	\captionsetup{font=small}
	\caption{{\small Results on \text{DAVIS2016}$_{val}$\!~\cite{perazzi2016benchmark}  (\S\ref{sec:ecs}).}}
	\label{table:davis16}
	\vspace{-12pt}
\end{table}

\begin{table}
	\centering
	\small
	\resizebox{1.00\columnwidth}{!}{
		\setlength\tabcolsep{6pt}
		\renewcommand\arraystretch{1.1}
		\begin{tabular}{lcccccc}
			 \hline\thickhline
		    \rowcolor{mygray}
		     & & & \multicolumn{2}{c}{Seen} & \multicolumn{2}{c}{Unseen} \\ \cline{4-7}
            \rowcolor{mygray}
            \multirow{-2}{*}{Method} & \multirow{-2}{*}{Synthetic} & \multirow{-2}{*}{Overall} & $\mathcal{J}$ $\uparrow$ & $\mathcal{F}$ $\uparrow$ & $\mathcal{J}$ $\uparrow$ & $\mathcal{F}$ $\uparrow$   \\  \hline \hline
         \multicolumn{7}{c}{\textit{YouTube-VOS2018 validation split}} \\ \hline
			SSTVOS\!~\cite{duke2021sstvos} & \xmark & 81.7 & 81.2 & - & 76.0 & -   \\
			CFBI+\!~\cite{yang2021collaborative}  & \xmark & 82.8 & 81.8 & 86.6 & 77.1 & 85.6  \\
			Joint\!~\cite{mao2021joint}  & \xmark  & 83.1 & 81.5 & 85.9 & 78.7 & 86.5  \\
			STCN\!~\cite{cheng2021rethinking}	& \xmark & 81.2  & 81.0 & 85.6 & 74.8 & 83.7   \\
			STCN+\textbf{\texttt{Ours}}  & \xmark  & \textbf{83.6} & \textbf{82.1} & \textbf{87.0} & \textbf{78.5} & \textbf{86.7} \\  \cdashline{1-7}[1pt/1pt]	
			XMem\!~\cite{cheng2022xmem}  & \xmark & 84.3 & 83.9 & 88.8 & 77.7 & 86.7  \\
			XMem+\textbf{\texttt{Ours}}  & \xmark  &  \textbf{85.6}  & \textbf{84.9} & \textbf{89.7} & \textbf{79.0} & \textbf{87.8}   \\
			\hline
			STM\!~\cite{oh2019video}  & \cmark  & 79.4 & 79.7 & 84.2 & 72.8 & 80.9  \\
			EGMN\!~\cite{lu2020video}  & \cmark  & 80.2 & 80.7 & 85.1 & 74.0 & 80.9  \\
			RMNet\!~\cite{xie2021efficient}  & \cmark  & 81.5 & 82.1 & 85.7 & 75.7 & 82.4  \\
			LCM\!~\cite{hu2021learning}  & \cmark  & 82.0 & 82.2 & 86.7 & 75.7 & 83.4   \\
			HMMN\!~\cite{seong2021hierarchical}  & \cmark & 82.6 & 82.1 & 87.0 & 76.8 & 84.6 \\
			AOT\!~\cite{yang2021associating}  & \cmark  & 84.1 & 83.7 & 88.5 & 78.1 & 86.1  \\	
			PCVOS\!~\cite{park2022per}  & \cmark & 84.6 & 83.0 & 88.0 & 79.6 & 87.9  \\
			\hline
			STCN\!~\cite{cheng2021rethinking}  &  & 83.0 & 81.9 & 86.5 & 77.9 & 85.7  \\
			STCN+\textbf{\texttt{Ours}}  & \multirow{-2}{*}{\cmark}  & \textbf{85.3} & \textbf{83.9} & \textbf{88.9} & \textbf{79.9} & \textbf{88.3} \\ \cdashline{1-7}[1pt/1pt]	
			XMem\!~\cite{cheng2022xmem}  &  & 85.7 & 84.6 & 89.3 & 80.2 & 88.7 \\
			XMem+\textbf{\texttt{Ours}}  & \multirow{-2}{*}{\cmark}  &  \textbf{86.9} & \textbf{85.5} & \textbf{90.2} & \textbf{81.6} & \textbf{90.4}  	\\	\hline
		 \hline
         \multicolumn{7}{c}{\textit{YouTube-VOS2019 validation split}} \\ \hline
		 KMN\!~\cite{seong2020kernelized}  & \cmark & 80.0 & 80.4 & 84.5 & 73.8 & 81.4  \\
		 LWL\!~\cite{bhat2020learning}  & \cmark & 81.0 & 79.6 & 83.8 & 76.4 & 84.2  \\
		 SSTVOS\!~\cite{duke2021sstvos}  & \cmark & 81.8 & 80.9 & - & 76.6 & -  \\
		 RDE\!~\cite{li2022recurrent}  & \cmark & 81.9 & 81.1 & 85.5 & 76.2 & 84.8  \\
		 HMMN\!~\cite{seong2021hierarchical}  & \cmark & 82.5 & 81.7 & 86.1 & 77.3 & 85.0  \\
         AOT\!~\cite{yang2021associating}  & \cmark  & 84.1 & 83.5 & 88.1 & 78.4 & 86.3   \\
         PCVOS\!~\cite{park2022per}  & \cmark & 84.6 & 82.6 & 87.3 & 80.0 & 88.3  \\
		 \hline
		 STCN\!~\cite{cheng2021rethinking}  &  & 82.7 & 81.1 & 85.4 & 78.2 & 85.9  \\
		 STCN+\textbf{\texttt{Ours}} & \multirow{-2}{*}{\cmark} &  \textbf{85.0}  & \textbf{83.2} & \textbf{87.7} & \textbf{80.6} & \textbf{88.6}  \\	\cdashline{1-7}[1pt/1pt]	
         XMem\!~\cite{cheng2022xmem}  &  & 85.5 & 84.3 & 88.6 & 80.3 & 88.6   \\
		 XMem+\textbf{\texttt{Ours}}  & \multirow{-2}{*}{\cmark} & \textbf{86.6} & \textbf{85.3} & \textbf{89.8} & \textbf{81.4} & \textbf{89.8}	\\
			\hline
		\end{tabular}
	}
    \captionsetup{font=small}
	\caption{{\small Results on YouTube-VOS2018$_{val}$\!~\&\!~2019$_{val}$$_{\!}$~\cite{xu2018youtube}  (\S\ref{sec:ecs}).}}
	\label{table:youtube}
	\vspace{-12pt}
\end{table}

\noindent\textbf{DAVIS2017$_{\!}$~\cite{pont2017davis}.} Table~$_{\!}$\ref{table:davis} reports the comparison results on DAVIS2017$_{val}$\&2017$_{test-dev}$. Our approach yields impressive results. Specifically, without synthetic video pre-training, our approach boosts the performance of STCN$_{\!}$~\cite{cheng2021rethinking} by a solid margin$_{\!}$~(\ie, 82.5\% $\rightarrow$ \textbf{84.7\%} on $\texttt{val}$, 73.9\% $\rightarrow$ \textbf{77.3\%}$_{\!}$ on$_{\!}$ $\texttt{test-dev}$),$_{\!}$ in$_{\!}$ terms$_{\!}$ of$_{\!}$ $\mathcal{J}\&\mathcal{F}_m$.$_{\!}$ Similarly,$_{\!}$~our approach improves the $\mathcal{J}\&\mathcal{F}_{m\!}$ of XMem$_{\!}$~\cite{cheng2022xmem} by \textbf{1.6\%} and \textbf{1.2\%}, on $\texttt{val}$ and $\texttt{test-dev}$, respectively. With the aid of synthetic video data, our approach based on STCN defeats all the competitors. Notably, on the top of XMem, our approach further pushes the state-of-the-arts forward, \ie, \textbf{87.7\%} $\mathcal{J}\&\mathcal{F}_m$ on $\texttt{val}$ and \textbf{82.0\%} $\mathcal{J}\&\mathcal{F}_m$ on $\texttt{test-dev}$.

\begin{figure*}[t]
	\begin{center}
		\includegraphics[width=\linewidth]{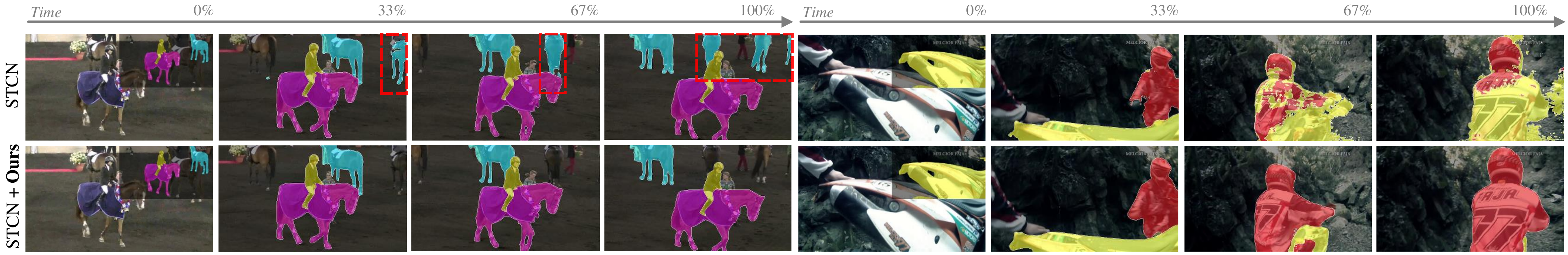}
		\put(-494.7,54.8){\tiny \rotatebox{90}{\cite{cheng2021rethinking}}}
	\end{center}
	\vspace{-18pt}
	\captionsetup{font=small}
	\caption{\small Qualitative results on YouTube-VOS2018$_{val\!}$~\cite{xu2018youtube} (\S\ref{sec:ecs}). The initial mask is presented in the upper right corner of the first frame.}
	\label{fig:results_youtube}
	\vspace{-10pt}
\end{figure*}

\noindent\textbf{YouTube-VOS$_{\!}$~\cite{xu2018youtube}.} Table$_{\!}$~\ref{table:youtube} compares our method against several top-leading approaches on YouTube-VOS2018$_{val}$\! \&\! 2019$_{val}$.$_{\!\!}$ As$_{\!}$ seen,$_{\!}$ our$_{\!}$ method$_{\!}$ greatly$_{\!}$ outperforms$_{\!}$~base
models, \ie, STCN: \textbf{83.6\%} \vs 81.2\%, and XMem: \textbf{85.6\%}~\vs 84.3\%,$_{\!}$  on$_{\!}$ $2018_{val\!}$ without$_{\!}$ synthetic$_{\!}$ data.$_{\!}$ With$_{\!}$~synthetic video pre-training, our approach respectively brings STCN and XMem to \textbf{85.3\%} and \textbf{86.9\%} on $2018_{val}$, as well as \textbf{85.0\%} and \textbf{86.6\%} on $2019_{val}$, setting new state-of-the-arts.

\noindent\textbf{Qualitative Results.} Fig.$_{\!}$~\ref{fig:results_youtube} displays qualitative comparison results on YouTube-VOS2018$_{val}$. We can observe that, compared with the original STCN, our approach generates more stable and accurate mask-tracking results, even on challenging scenarios with fast motion or occlusion.

\subsection{Diagnostic Experiment}\label{sec:ede}
For thorough evaluation, we conduct ablation studies on DAVIS2017$_{val}$$_{\!}$~\cite{pont2017davis} and YouTube-VOS2018$_{val}$$_{\!}$~\cite{xu2018youtube}.

\noindent\textbf{Component-wise$_{\!}$ Analysis.$_{\!}$} We$_{\!}$ first$_{\!}$ ablate$_{\!}$ the$_{\!}$ effects$_{\!}$ of$_{\!}$ our core algorithm components, \ie, pixel-level correspondence\\
\noindent learning$_{\!}$ ($\mathcal{L}_{\text{PCL}}$,$_{\!}$ Eq.$_{\!}$~\ref{eq:7})$_{\!}$ and$_{\!}$ object-level$_{\!}$ correspondence$_{\!}$ lear- ning ($\mathcal{L}_{\text{OCL}}$, Eq.$_{\!}$~\ref{eq:11}). In Table$_{\!}$~\ref{table:ablation1}, \#1~row gives the results~of the baseline models, \ie, STCN$_{\!}$~\cite{cheng2021rethinking} and XMem$_{\!}$~\cite{cheng2022xmem}; \#2 and \#3 rows list the performance obtained by additionally considering$_{\!}$ $\mathcal{L}_{\text{PCL\!}}$ and$_{\!}$ $\mathcal{L}_{\text{OCL\!}}$ individually;$_{\!}$ \#4$_{\!}$ row$_{\!}$ provides the$_{\!}$~scores$_{\!}$ of$_{\!}$ our$_{\!}$ full$_{\!}$ algorithm.$_{\!}$ Comparisons$_{\!}$ between$_{\!}$~base- lines (\#1 row) and variants with single component bonus (\#2 row and \#3 row) verify the efficacy of each module$_{\!}$ design.$_{\!}$ When$_{\!}$ comprehensively$_{\!}$ comparing$_{\!}$ the$_{\!}$ four$_{\!}$ rows, we can find that the  best performance is acquired after combining the two components (\ie, \#4 row). This suggests that the two components can cooperate harmoniously and confirms the {joint} effectiveness of our overall algorithmic design.

\noindent\textbf{Training Speed.} {Training speeds are also compared} in Table$_{\!}$~\ref{table:ablation1}. As seen, our algorithm only introduces negligible delay in the training speed (about 7\%$\sim$8\%), {while leveraging such a performance leap}. More essentially, {the adaption of our training mechanism does not affect the original inference process in both complexity and efficiency.}

\begin{table}
	\renewcommand\thetable{4}
   \centering
   \small
   \resizebox{\columnwidth}{!}{
      \tablestyle{3pt}{1.05}
      \begin{tabular}{c|cc||cc|c||cc|c}
        \hline\thickhline
           \rowcolor{mygray}
          	&$\mathcal{L}_{\text{PCL}}$ &$\mathcal{L}_{\text{OCL}}$ &  \multicolumn{3}{c||}{STCN$_{\!}$~\cite{cheng2021rethinking}} & \multicolumn{3}{c}{XMem$_{\!}$~\cite{cheng2022xmem}} \\ \cline{4-9}
            \rowcolor{mygray}
			\multirow{-2}{*}{\#} & \multirow{1}{*}{(Eq.~\ref{eq:7})} & \multirow{1}{*}{(Eq.~\ref{eq:11})}  & $D_{17}$  & $Y_{18}$ &min/epoch  & $D_{17}$ &  $Y_{18}$  &min/epoch\\ \hline\hline
            1 &  &  &   85.4 & 83.0 & 2.29  & 86.2	&	85.7 & 3.42\\  \hline
            2 & \cmark &  & 86.3&  84.8 & 2.43 & 87.2&86.6 &3.59\\
            3 & & \cmark &  85.9 &  84.0 & 2.39 &  86.9& 86.3&3.53 \\\hline
            4 & \cmark&\cmark  &  86.6&  85.3  & 2.50 & 87.7& 86.9 & 3.66  \\
			\hline
      \end{tabular}
   }
   \captionsetup{font=small}
   \caption{\small Analysis of essential components of our~training algorithm on DAVIS2017$_{val}$ ($D_{17}$)$_{\!}$~\cite{pont2017davis} and YouTube-VOS2018$_{val\!}$ ($Y_{18}$)$_{\!}$~\cite{xu2018youtube} (\S\ref{sec:ede}). Training speed is reported in min/epoch.}
   \label{table:ablation1}
   \vspace{-6pt}
\end{table}

\begin{table}
	\centering
	\small
\resizebox{0.95\columnwidth}{!}{
      \tablestyle{6pt}{1.05}
      \begin{tabular}{c|cc||cc||cc}
        \hline\thickhline
           \rowcolor{mygray}
          	&\multicolumn{2}{c||}{Negative Sample} &  \multicolumn{2}{c||}{STCN$_{\!}$~\cite{cheng2021rethinking}} & \multicolumn{2}{c}{XMem$_{\!}$~\cite{cheng2022xmem}} \\ \cline{2-7}
            \rowcolor{mygray}
			\multirow{-2}{*}{\#} &\textit{inter}-video &\textit{intra}-video  & $D_{17}$  & $Y_{18}$ & $D_{17}$ &  $Y_{18}$ \\ \hline\hline
            1 &  &  & 85.4 & 83.0  & 86.2	&	85.7 \\  \hline
            2 & \cmark & & 86.1 & 84.5 & 86.9  & 86.4\\
            3 & & \cmark & 85.8  &  83.9 &  86.6 & 86.2\\\hline
            4 & \cmark&\cmark  &  86.3&  84.8  & 87.2& 86.6   \\
			\hline
      \end{tabular}
   }
	\captionsetup{font=small}
	\caption{\small Comparison$_{\!}$ of$_{\!}$ different$_{\!}$ strategies$_{\!}$ (\S\ref{sec:ede})$_{\!}$ for$_{\!}$ sampling$_{\!}$ neg-$_{\!}$ ative$_{\!}$ pairs$_{\!}$ during$_{\!}$ pixel-level$_{\!}$ correspondence$_{\!}$ learning$_{\!}$ ($\mathcal{L}_{\text{PCL}}$,$_{\!}$ Eq.$_{\!}$~\ref{eq:7}).$_{\!}$}
	\label{table:contrast}
	\vspace{-8pt}
\end{table}

\noindent\textbf{Negative Pair Sampling for Pixel-level Correspondence Learning.} During the computation of our pixel-level con- sistency$_{\!}$ based$_{\!}$ contrastive$_{\!}$ correspondence$_{\!}$ learning$_{\!}$ objective$_{\!}$ $\mathcal{L}_{\text{PCL}}$ (\textit{cf}.$_{\!}$~Eq.$_{\!}$~\ref{eq:7}), we sample non-corresponding pairs from both the current training video as well as other videos within the same training batch as negative examples. Next, we investigate the impact of such negative example sampling strategy. As {indicated} by Table$_{\!}$~\ref{table:contrast}, exploring both inter- and inter-video negative correspondence leads to the best performance (\ie, \#4 row). This is because the performance of contrastive correspondence learning heavily relies on the diversity (or quality) of the negative samples.

\noindent\textbf{Object Source for Object-level Correspondence$_{\!}$ Learning.} During$_{\!}$ the$_{\!}$ computation$_{\!}$ of$_{\!}$ our$_{\!}$ object-level consistency$_{\!}$ based$_{\!}$ contrastive$_{\!}$ correspondence$_{\!}$ learning objective$_{\!}$ $\mathcal{L}_{\text{OCL}}$ (\textit{cf}.$_{\!}$~Eq.$_{\!}$~\ref{eq:11}), we adopt Selective Search$_{\!}$~\cite{uijlings2013selective} to automatically\\ \noindent generate a large set of potential object candidates, instead of only using a few annotated object instances -- only occupying a small ratio of the objects in the training videos. Table$_{\!}$~\ref{table:proposal search} studies the influence of different sources of object proposals. It can be found that the best performance is achieved$_{\!}$ by$_{\!}$ exploring$_{\!}$ both$_{\!}$ manually-labeled$_{\!}$ object$_{\!}$ instances as well as massive automatically-mined object proposals as training samples. This is because the considerable number of object proposals can improve the richness of training samples, enabling robust correspondence matching.

\begin{table}
	\centering
	\small
\resizebox{\columnwidth}{!}{
      \tablestyle{4pt}{1.05}
      \begin{tabular}{c|cc||cc||cc}
        \hline\thickhline
           \rowcolor{mygray}
          	&\multicolumn{2}{c||}{Object Source} &  \multicolumn{2}{c||}{STCN$_{\!}$~\cite{cheng2021rethinking}} & \multicolumn{2}{c}{XMem$_{\!}$~\cite{cheng2022xmem}} \\ \cline{2-7}
            \rowcolor{mygray}
			\multirow{-2}{*}{\#} &\textit{manu}. annotated &\textit{auto}. discovered  & $D_{17}$  & $Y_{18}$ & $D_{17}$ &  $Y_{18}$ \\ \hline\hline
            1 &  &  & 85.4 & 83.0  & 86.2	&	85.7 \\  \hline
            2 & \cmark & & 85.7 & 83.5&  86.5 & 86.1\\
            3 & & \cmark &  85.8 &  83.7 & 86.7 & 86.0\\\hline
            4 & \cmark&\cmark  &  85.9&  84.0  & 86.9& 86.3  \\
			\hline
      \end{tabular}
   }
	\captionsetup{font=small}
	\caption{Comparison$_{\!}$ of$_{\!}$ different$_{\!}$ sources$_{\!}$ of$_{\!}$ objects$_{\!}$ (\S\ref{sec:ede}), for object-level$_{\!}$ correspondence$_{\!}$ learning$_{\!}$ ($\mathcal{L}_{\text{OCL}}$,$_{\!}$ Eq.$_{\!}$~\ref{eq:11}).}
	\label{table:proposal search}
	\vspace{-8pt}
\end{table}

\noindent\textbf{Can$_{\!}$ VOS$_{\!}$ Benefit$_{\!}$ from$_{\!}$ Existing$_{\!}$ Self-supervised$_{\!}$ Corres- pondence$_{\!}$ Learning$_{\!}$ Techniques?$_{\!}$}} One$_{\!}$ may$_{\!}$ be$_{\!}$ interested$_{\!}$ in if$_{\!}$ VOS$_{\!}$ can$_{\!}$ be$_{\!}$ boosted$_{\!}$ by$_{\!}$ existing$_{\!}$ correspondence$_{\!}$ learning techniques.$_{\!}$  We$_{\!}$ select$_{\!}$ three$_{\!}$ representative$_{\!}$ top-leading$_{\!}$ cor- respondence$_{\!}$ algorithms:$_{\!}$ reconstruction$_{\!}$ based\!~\cite{li2022locality},$_{\!}$ cycle-consistency$_{\!}$ based\!~\cite{jabri2020space},$_{\!}$  and$_{\!}$ contrastive$_{\!}$ learning$_{\!}$ based\!~\cite{araslanov2021dense},$_{\!}$ and$_{\!}$~apply$_{\!}$ them$_{\!}$ to$_{\!}$ STCN\!~\cite{cheng2021rethinking}.$_{\!}$ The$_{\!}$ results$_{\!}$ are$_{\!}$ summarized$_{\!}$~in {Table}$_{\!}$~\ref{table:ssl}.$_{\!}$ As$_{\!}$ seen,$_{\!}$ little$_{\!}$ or$_{\!}$ even$_{\!}$ negative$_{\!}$ performance$_{\!}$ gain$_{\!}$~is obtained.$_{\!}$ It$_{\!}$ is$_{\!}$  possibly$_{\!}$ because:$_{\!}$ \textbf{i)}$_{\!}$ none$_{\!}$ of$_{\!}$ them$_{\!}$ adopts$_{\!}$ object-level$_{\!}$ matching,$_{\!}$ but$_{\!}$ VOS$_{\!}$ is$_{\!}$ \textit{object-aware};$_{\!}$ \textbf{ii)}$_{\!}$ their$_{\!}$ training strategy$_{\!}$ is$_{\!}$ relatively$_{\!}$ simple$_{\!}$ (\eg,$_{\!}$ color$_{\!}$ space$_{\!}$ reconstruction$_{\!}$ \cite{li2022locality},$_{\!}$
\textit{within}-video$_{\!}$ cycle-tracking\!~\cite{jabri2020space}),$_{\!}$ tending$_{\!}$ to$_{\!}$ overem-
phasize$_{\!}$ low-level$_{\!}$ cues;$_{\!}$ \textbf{iii)}$_{\!}$ \cite{xu2021rethinking}$_{\!}$ learns$_{\!}$ correspondence$_{\!}$ based$_{\!}$
 on$_{\!}$  frame-level$_{\!}$ similarity,$_{\!}$  which$_{\!}$  may$_{\!}$  be$_{\!}$ sub-optimal$_{\!}$ for$_{\!}$ the$_{\!}$
 scenarios$_{\!}$ involving$_{\!}$ multiple$_{\!}$ objects.$_{\!}$ In$_{\!}$ contrast,$_{\!}$  our$_{\!}$ method
 incorporates pixel- and object-level correspondence learning simultaneously, facilitating$_{\!}$ more$_{\!}$ VOS-aligned$_{\!}$ training.



\begin{table}
    \centering
    \small
    \resizebox{\columnwidth}{!}{
     \tablestyle{2pt}{1.1}
     \begin{tabular}{c||c||cccc}
         \hline\thickhline
         \cellcolor[gray]{0.9} Dataset &VOS model\!~\cite{cheng2021rethinking} & LIIR\!~\cite{li2022locality} & CRW\!~\cite{jabri2020space} & VFS\!~\cite{xu2021rethinking} & \textbf{Ours} \\ \hline \hline
         \cellcolor[gray]{0.9} DAVIS2017$_{val\!}$ & 85.4  & 85.6 & 85.5 & 85.8 &   \textbf{86.6}   \\
         \cellcolor[gray]{0.9} YouTube2018$_{val\!}$ & 83.0 & 83.4 & 83.3  & 83.9 &   \textbf{85.3} \\
         \hline
     \end{tabular}
    }
    \captionsetup{font=small}
    \caption{\small Comparison$_{\!}$ with$_{\!}$ self-supervised$_{\!}$ corresponding$_{\!}$ learning$_{\!}$ methods$_{\!}$ on$_{\!}$ DAVIS2017$_{val}$$_{\!}$~\cite{pont2017davis} and$_{\!}$  YouTube-VOS2018$_{val\!}$$_{\!}$~\cite{xu2018youtube}$_{\!}$.}
    \label{table:ssl}
    \vspace{-8pt}
\end{table}

\section{Conclusion}
Recent efforts in matching-based VOS are dedicated to more powerful model designs, while neglecting the absence of explicit supervision signals for space-time correspondence matching, which yet is the heart of the whole system. Noticing this, we take the lead in incorporating self-constrained correspondence training target with matching-based VOS models, begetting a new training {mechanism} for fully-supervised VOS and boosting excellent performance in a portable manner. It enjoys several charms: \textbf{i)} no modification on network architecture, \textbf{ii)} no extra annotation budget, and \textbf{iii)} no inference {time delay and efficiency burden}.



{\small
\bibliographystyle{ieee_fullname}
\bibliography{egbib}
}

\end{document}